\documentclass[11pt]{article}
\usepackage[round,authoryear]{natbib}
\PassOptionsToPackage{no-math}{fontspec}
\usepackage[UTF8,scheme=plain,fontset=fandol]{ctex}
\usepackage{metacircle}
\usepackage{multirow}
\usepackage{adjustbox}
\usepackage{float}
\setcitestyle{authoryear,round,citesep={;},aysep={,},yysep={;}}

\title{Why Do Conventional World Models Fail to Learn Cellular Automata?}
\author{
  Shaoyang Guo$^{1,2,*}$\thanks{Contact: guoshaoyang@stu.pku.edu.cn} and Ziming Liu$^{1,3,\dagger}$\\
  \smallskip
  \small $^1$Meta Circle(元环智能) \quad $^2$Peking University \quad $^3$Tsinghua University\\
  \small $^*$Core contributor \quad $^\dagger$Corresponding author
}
\date{}
\paperdate{September 26, 2026}
\papertype{Research Paper}
\projecturl{https://github.com/guoshaoyang-pku/momentum-induction/tree/release}

\begin{document}

\newif\ifarchbadges
\archbadgestrue
\newcommand{\arch}[1]{\ifarchbadges\,\raisebox{-0.32ex}{\includegraphics[height=1.75ex]{figs/badges/#1.pdf}}\fi}
\definecolor{insightink}{HTML}{37474F}
\newtcolorbox{insight}[1]{enhanced, colback=insightink!4!white, colframe=insightink!50!white,
  colbacktitle=insightink!9!white, coltitle=insightink, fonttitle=\bfseries, title={#1},
  arc=1.0mm, boxrule=0.45pt, left=5pt, right=5pt, top=3pt, bottom=3pt,
  toptitle=1.5pt, bottomtitle=1.5pt, before skip=4pt, after skip=4pt}

\setcounter{footnote}{2}
\maketitle

\begin{metaabstract}
Although conventional world models --- auto-regressive or diffusion models based on transformers or convolutional networks --- may learn surface statistics of world dynamics, can they learn the exact world dynamics from its observed history? Leveraging cellular automata as a simple testbed, we find the answer to be \textbf{no} in many cases. Conventional architectures predict most pixels correctly yet rarely complete a rollout: a CNN predicts 96.3\% of cells but completes 18.9\% of rollouts; a joint diffusion model completes none. We trace the gap to three failure modes of these world models --- namely, they fail to exactly capture spatial locality, temporal locality or temporal stability. Simple changes repair each: (1) for \textbf{spatial locality}, two-dimensional rotary positions lift a transformer from 39.1\% to 100\% on the Game of Life; (2) for \textbf{temporal locality}, handing each token its cell's previous-frame neighbourhood lifts the same transformer from 25.8\% to 99.9\% on unseen rules; (3) for \textbf{temporal stability}, causal freezing lifts the same diffusion weights from 42.2\% to 99.9\%. None of the three changes touches the architectural backbone; each only modifies the information flow within it. We also compare joint and ordered sampling on billiards and, in an exploratory study, on a simulated Burgers equation.
\end{metaabstract}

\begin{tcolorbox}[
  enhanced, breakable,
  colback=white, colframe=MetaViolet,
  boxrule=0.6pt, arc=3mm,
  left=3.5mm, right=3.5mm, top=2.5mm, bottom=2.5mm,
  before skip=9pt, after skip=9pt
]
\noindent{\color{MetaPurple}\bfseries Acknowledgements.}\;\;
We thank \textbf{Qingyu Qu} and \textbf{Chencheng Tang} (Tsinghua University) for their important guidance on the research direction of this work.
\end{tcolorbox}

\section{Introduction}
\label{sec:intro}

World models aim to learn the rules governing a world from observations and use them to predict how the world evolves \citep{ha2018world, lecun2022path}. Existing work has explored a range of objectives, including learning physical dynamics \citep{kang2025videolaw}, predicting future states \citep{hafner2023mastering}, maintaining scene consistency across time and viewpoints \citep{openai2024sora}, and capturing the interactions among objects and agents \citep{battaglia2018relational, micheli2022transformers}. However, generative models often predict well without recovering the world model behind the data \citep{vafa2024evaluating, vafa2025foundation, kang2025videolaw}, and the failure is hard to analyse because the rule itself is unknown. Cellular automata (CA, Figure~\ref{fig:infoflow}) provide a simple yet expressive setting for addressing this question. Wolfram's studies of cellular automata showed how simple local rules can generate rich and complex behaviour \citep{wolfram1983statistical}. In a CA, each cell updates according to its own state and those of its neighbours; in Conway's Game of Life, such local interactions produce patterns that move, interact, and collide \citep{gardner1970fantastic}. This combination of simple, well-defined rules and rich behaviour makes cellular automata a clean testbed for investigating why standard world models fail to infer underlying world rules.

\begin{insight}{Research question}
Why can't standard world models --- autoregressive or diffusion models based on transformers or convolutional networks --- infer an unseen cellular automaton's rule from observed history?
\end{insight}

\begin{figure}[t]
\centering
\includegraphics[width=\linewidth]{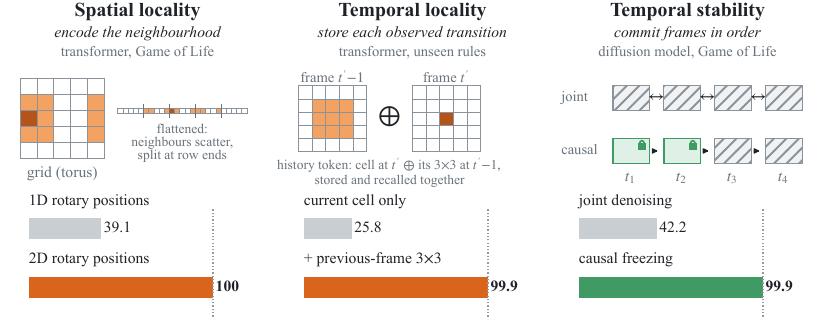}
\caption{\textbf{Three failures of standard models, and the change that repairs each} (bars: exact rollouts, \%). Each panel isolates one property; the grey boxes below state the failure and its fix, and Sections~\ref{sec:spatial}--\ref{sec:stability} give the evidence. \emph{Left, spatial locality}: a transformer that reads the grid as a flat token sequence must know which tokens are neighbours; raster order scatters a cell's neighbourhood and splits it where rows wrap around the torus, which one-dimensional positions cannot resolve and two-dimensional positions can. \emph{Middle, temporal locality}: on an unseen rule the model must pair each situation with its outcome one frame later; giving every history token its cell's previous-frame $3{\times}3$ neighbourhood stores each observed transition in one place, so situation and outcome are recalled together. \emph{Right, temporal stability}: under joint denoising every frame stays open to revision by the others until the end; causal freezing, with the same weights, locks each frame before the next is built on it.}
\label{fig:teaser}
\end{figure}

Several studies have investigated predicting future states of cellular automata from observations \citep{gilpin2019cellular, springer2021hard, burtsev2024learning, berkovich2025lifegpt, berkovich2026automatagpt}. However, these works have primarily focused on one-step prediction or pixel-level accuracy, rather than whether a model can generate a correct long-horizon continuation (rollout) of the system. For example, a model can achieve $99.8\%$ pixel accuracy while maintaining a correct rollout for only $64\%$ of the evaluated trajectories (Figure~\ref{fig:twometrics}). Here, we evaluate standard architectures, including convolutional neural networks, transformers, and diffusion models, on cellular automata using rollout success as a metric for long-horizon prediction. Surprisingly, we find that these architectures fail to reliably complete the task, despite their high short-term prediction accuracy. This motivates us to investigate why current architectures struggle with this seemingly simple task and how they can be improved. We identify a set of simple yet general inductive biases that are absent from current architectures; incorporating these biases enables the models to continue cellular automata exactly over long horizons.

The three inductive biases we identify are as follows (Figure~\ref{fig:teaser}):

\begin{insight}{Spatial locality --- step 1, seeing the situation (Section~\ref*{sec:spatial})}
\textbf{Failure mode:} flattening the grid into one token sequence hides which cells are neighbours. With one-dimensional rotary positions, a transformer executes the known Game of Life in only 39.1\% of rollouts, and its errors gather on the wrap-around edge.\\
\textbf{Solution:} two-dimensional positions, with the parameters unchanged: 100\%.
\end{insight}

\begin{insight}{Temporal locality --- steps 2--3, matching a situation to what followed it (Section~\ref*{sec:temporal})}
\textbf{Failure mode:} a situation and its outcome lie one frame apart, and standard models rarely learn to bind them: the standard CNN completes 18.9\% of unseen-rule rollouts; no standard transformer within our budget is exact.\\
\textbf{Solution:} store each situation with its successor---a two-frame convolution, a KV shift \citep{xu2024kvshift}, or the previous-frame neighbourhood on each token, which lifts the same transformer from 25.8\% to 99.9\% (Figure~\ref{fig:twoinductions}).
\end{insight}

\begin{insight}{Temporal stability --- the rollout loop, settling each frame before it is used (Section~\ref*{sec:stability})}
\textbf{Failure mode:} joint denoising computes later frames from earlier ones that are never settled; more network calls and deeper denoisers do not fix it.\\
\textbf{Solution:} causal freezing \citep{chen2024diffusion}---denoise the frames in order, keeping each as clean context: with weights unchanged, a depth-8 plain denoiser rises from 42.2\% to 99.9\% on the Game of Life, while billiards barely changes (Figure~\ref{fig:billiards}).
\end{insight}

Each section asks the same three questions of one step: is the needed information reachable, does training learn to use it, and which small change supplies the missing link? Reachability is checked before training (Section~\ref{sec:task}): in the CNN and the transformer the evidence was within reach---inside the receptive field or the context---and what failed was learning to use it, whereas the standard denoiser's four-layer field does not even reach the oldest observed frames. Each comparison changes one link and holds the rest of the model fixed. None of the devices is new---the KV shift comes from language models, causal freezing from video diffusion; what the automaton adds is a world in which the reason each one is needed can be stated in advance and tested. Related work is discussed in Appendix~\ref{app:related}.

\section{Cellular automata as a testbed}
\label{sec:task}

\begin{figure}[t]
\centering
\begingroup
\makeatletter
\long\def\@makecaption#1#2{%
  \vskip 6pt
  {\fontsize{9}{10.5}\selectfont\setlength{\parskip}{0pt}%
   \raggedright\noindent #1: #2\par}}
\makeatother
\begin{minipage}[t]{0.57\linewidth}
\vspace{0pt}
\includegraphics[width=\linewidth]{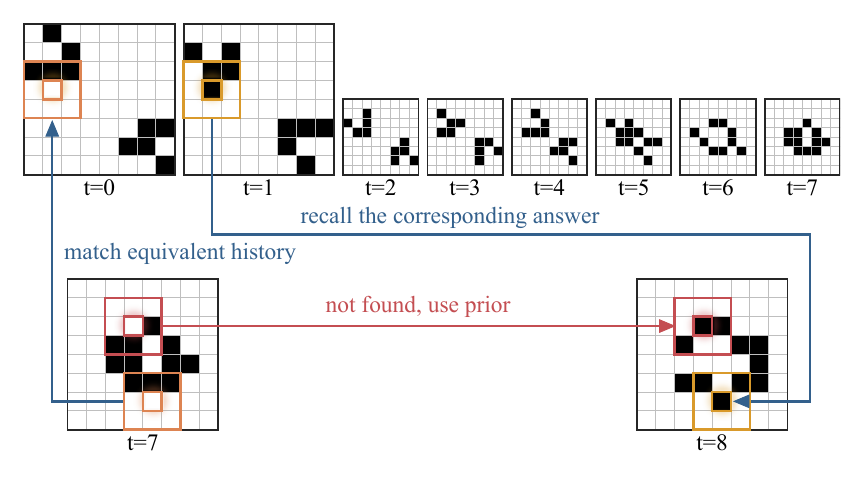}
\caption[The information flow required for prediction.]{\textbf{The information flow required for prediction.} The grid is observed at $t=0,\dots,7$; the task is to predict the state of a queried cell at $t=8$.
\par\noindent\textbf{Step 1: identify the current situation.} At $t=7$, record the queried cell's state and the number of black cells among its eight neighbours (orange box).
\par\noindent\textbf{Step 2: match a past situation} (blue). Find an earlier observed transition with the same central state and neighbour count; the match shown occurs at $t=0$.
\par\noindent\textbf{Step 3: recall its outcome} (blue). Read the matched cell's state in the following frame ($t=1$) and copy that bit to the queried cell at $t=8$; the same fixed rule maps identical situations to identical outcomes.
\par\noindent\textbf{No match} (red). Direct retrieval cannot determine the next state, so prediction relies on prior knowledge of the rule family; the L2 and L3 test corpora contain no such cells, and L4 tests them separately. The $8\times 8$ torus has periodic boundaries (edges wrap).}
\label{fig:infoflow}
\end{minipage}\hfill
\begin{minipage}[t]{0.40\linewidth}
\vspace{0pt}
\begingroup
\makeatletter\def\@captype{table}\makeatother
\centering
\footnotesize
\begin{tabular}{@{}ll@{}}
\toprule
\multicolumn{2}{@{}l@{}}{\textbf{Goal: see 8 frames, predict next 8 exactly}}\\
\midrule
L1 & a fixed rule (Game of Life)\\
L2 & seen worlds; context tells which\\
L3 & unseen worlds; context gives the rule\\
L4 & evidence withheld; a prior is needed\\
\bottomrule
\end{tabular}
\caption{The four evaluation levels.}
\label{tab:levels}
\endgroup
\vskip6pt
\includegraphics[width=\linewidth]{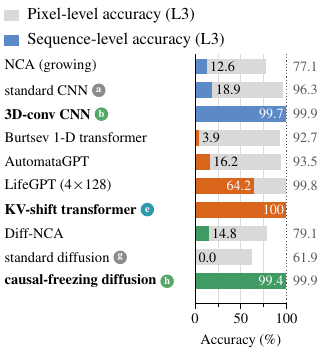}
\caption{\textbf{Two metrics on held-out rules.} Grey: pixel-level accuracy, the fraction of predicted cells correct. Coloured: SeqAcc, the fraction of rollouts with all $8\times64=512$ cells correct. Means over three seeds; per-seed values in Appendix~\ref{app:recipes}.}
\label{fig:twometrics}
\end{minipage}
\endgroup
\end{figure}

In the CA testbed, each world is a binary outer-totalistic cellular automaton on an $8\times 8$ torus. Its rule is an 18-row lookup table
\begin{equation}
r:\{0,1\}\times\{0,\dots,8\}\to\{0,1\},
\end{equation}
which maps a cell's own state $s$ and its number of black neighbours $n$ to its next state. We call $(s,n)$ the cell's \emph{situation}. There are $2^{18}=262{,}144$ rules, including the Game of Life. A fixed-seed permutation splits them into two disjoint halves of $131{,}072$ rules, one for training and one held out.

A model observes 8 frames and predicts the next 8, a \textbf{rollout}: autoregressive models generate it one frame at a time, feeding each prediction back as input, while a diffusion model denoises all eight frames jointly. 

Four levels grade the difficulty (Table~\ref{tab:levels}). L1 fixes one world, the Game of Life: can the model execute a known rule? L2 uses new trajectories of training-half worlds. L3, the headline level, uses held-out worlds, so the rule must be inferred from the eight observed frames alone. L4 builds on L3, but the context may not contain the answer: a queried situation may never appear among the observed transitions. The model must then use a prior we build into the rule family: two situations with opposite centre cells and the same neighbour count have opposite outcomes, $r(1,n)=1-r(0,n)$ for $n\in\{3,4\}$. So it looks for the centre-flipped situation in the context, which is guaranteed to be there, and negates what followed it.

The metric is strict sequence accuracy,
\begin{equation}
\mathrm{SeqAcc}=\Pr\!\left[\text{all } 8\times 64 = 512 \text{ predicted cells correct}\right].
\end{equation}
Test corpora are \emph{self-consistent}: every queried situation appears among the observed transitions, except at L4, which withholds it on purpose. Without this filter, a third of the trajectories query a situation that the observed frames never show, so the observations do not determine their continuation. Training streams are filtered the same way. Unless a caption says otherwise, numbers are final-step values without checkpoint selection, averaged over three seeds; the few exceptions---figure cells with fewer registered seeds, and the particle experiment of Appendix~\ref{app:bridge}, whose checkpoint is validation-selected---are marked where they appear. Per-seed values and training budgets are in Appendices~\ref{app:models} and~\ref{app:recipes}.

\textbf{Standard models.} We build three standard next-frame models as a language or video modeller would, and train them on identical data. The standard CNN\arch{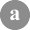} stacks the eight observed frames as input channels; its $9{\times}9$ receptive field covers the torus. The standard transformer\arch{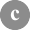} is a causal decoder over raster tokens: sixteen frames flattened into 1024 tokens with learned absolute positions, at four widths (0.17M--6.6M parameters, 2--8 layers). The standard diffusion model\arch{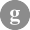} is a four-layer 3D-convolutional denoiser in which all eight predicted frames share one noise level. In the CNN and the transformer every observed transition can reach every prediction; the standard denoiser is narrower, its depth-4 temporal field reaching only the four most recent observed frames.

\begin{figure}[t]
\centering
\includegraphics[width=\linewidth]{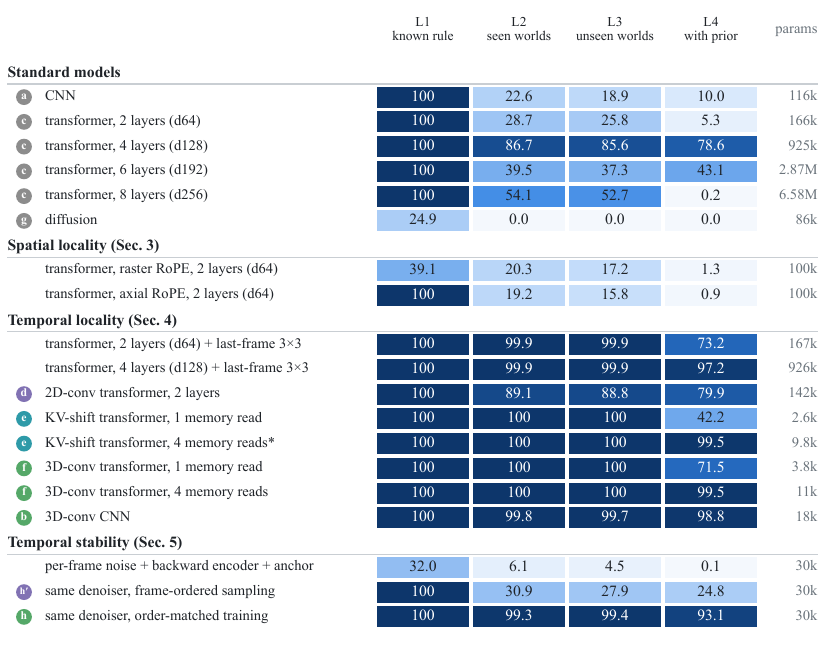}
\caption{\textbf{Main results.} SeqAcc (\%) at the four levels, mean over registered seeds (per-seed values in Figure~\ref{fig:mainseeds}). Rows are grouped by the property of Sections~\ref{sec:spatial}--\ref{sec:stability} that they test. Standard-transformer rungs tie depth to width (d64--d256, 2--8 layers, four heads). A ``memory read'' is one attention retrieval over the frozen evidence memory (Figure~\ref{fig:e21}); * marks an L4 cell trained with the evidence-balanced loss. The three denoiser rows share one architecture---per-frame noise levels, a backward-looking canvas encoder, and a clean anchor frame added to the standard denoiser---sampled jointly, sampled in frame order with the same weights (h$'$, causal freezing), or trained matched to that order (h). Coloured letters index the design-space panels (Figures~\ref{fig:dsconv}--\ref{fig:dsdiff}); protocol exceptions are in Table~\ref{tab:main}.}
\label{fig:main}
\end{figure}

\textbf{Scored per pixel, the metric of prior work, every baseline looks nearly solved; scored per sequence, all fail} (Figure~\ref{fig:twometrics}). Pixel-level accuracy \citep{berkovich2025lifegpt, berkovich2026automatagpt, burtsev2024learning} hides rule errors: at 99.9\% per pixel, independent errors would still spoil about two in five rollouts (Appendix~\ref{app:protocol}). Every baseline in Figure~\ref{fig:twometrics} with a one-step prediction path---ours, and those adapted from prior work within a predeclared budget of 0.5--1.5M parameters---gets at least 77\% of pixels right one step ahead, yet few rollouts right: the best, the LifeGPT of the introduction \citep{berkovich2025lifegpt}, completes only 64.2\% of held-out rollouts, and every other model shown in this comparison completes at most 18.9\%. The failures are not shared evenly: flattening the grid is the transformer's problem, joint denoising the diffusion model's, and binding a situation to its outcome everyone's. Figure~\ref{fig:main} lists our models, grouped by the step of Figure~\ref{fig:infoflow} that Sections~\ref{sec:spatial}--\ref{sec:stability} supply.

\section{Spatial locality: keep neighbours together}
\label{sec:spatial}

\textbf{With one-dimensional rotary positions, a transformer executes the Game of Life in 39.1\% of rollouts, and 95\% of its errors fall on the 28 border cells.} A cell's next state depends on its $3{\times}3$ neighbourhood. A convolution sees this neighbourhood directly, and the standard CNN executes the Game of Life exactly (L1 in Figure~\ref{fig:main}). A raster transformer instead reads each frame as 64 consecutive tokens: a cell's previous-frame neighbourhood becomes three runs of three tokens about 64 positions earlier, and at the edge of the torus each run wraps to the other side of its row. With learned absolute positions, the standard transformer learns this layout and is exact. With rotary position encoding (RoPE) along the raster sequence, as in most language models, the d64 transformer is not (Table~\ref{tab:e14}, Appendix~\ref{app:serialized}): at the wrap-around edge a fixed raster offset no longer points to a fixed neighbour, and that is where the errors sit.

\textbf{Two-dimensional positions, with the same parameters, give 100\%.} Rotating by row and column separately (axial 2D RoPE), with identical parameters and initialization, gives 100\% on every seed. Toroidal and deliberately mis-periodic variants also reach 99.4--100\%: what matters is two-dimensional position, not the exact period. On unseen rules with one token per cell, 2D rotary positions lift the d128 transformer from 82.6\% to 99.3\%, higher on every seed, and at d64 neither scheme exceeds 18\%; with the previous-frame neighbourhood added to each token, the preregistered 2D-over-1D gains missed their per-seed thresholds (Figure~\ref{fig:e22}, Appendix~\ref{app:serialized}).

\textbf{Attention can learn the layout; what it lacks is reliability.} With learned positions the same stack is exact on the known rule and reaches 85.6\% on unseen rules at d128---yet no width is exact within our budget (Appendix~\ref{app:scladder}). The previous-frame binding of Section~\ref{sec:temporal} makes the same decoders near-exact with a few hundred added parameters. The lesson is not that attention cannot represent locality but that it does not reliably learn it. Structure buys reliability, not capability---and reliability is the reason to build locality in rather than stack width.

\textbf{Spatial locality is necessary but not sufficient.} The standard CNN, local by construction, executes the Game of Life exactly yet fails most held-out rollouts; what it lacks is temporal locality.

\section{Temporal locality: bind each state to its successor}
\label{sec:temporal}

Induction needs each observed situation and the outcome it produced in one place that the current query can find. In the input the two lie a frame apart, and no standard model within our budget reliably learns to bring them together. We call models given this binding \emph{momentum induction} models: each observed situation is stored with its continuation, so recalling the situation carries the continuation with it. The models of the main text receive structure and initialization only, never extra supervision, apart from the L4 cells marked * in Figure~\ref{fig:main}, whose loss uses a training-side task label (Appendix~\ref{app:models}). \emph{Learned} models train their detectors from random initialization; \emph{analytic} models use hand-set detectors and train only the head.

\textbf{The standard CNN sees the whole torus and all eight frames, yet completes 18.9\% of held-out rollouts, and five times the training steps do not change this} (Appendix~\ref{app:recipes}). Convolutions can represent the rule exactly \citep{gilpin2019cellular}, but even the fixed Life rule is hard to learn \citep{springer2021hard}. The standard transformer leaves the marginal predictor, which ignores the evidence and scores 4.1\%, within its first 18k steps or never: 200k steps and a halved learning rate free no stuck run, and at the two widest models one seed in three stays stuck (Appendix~\ref{app:scladder}). Runs that escape infer rather than recall---L2 tracks L3 at every width---yet none is exact. This falsifies both of our preregistered predictions for this model: that it would stay near zero, and that accuracy would rise monotonically with width (Appendix~\ref{app:process}).

\subsection{Convolution: count, then look up}
\label{sec:conv}

\textbf{A convolution over two adjacent frames makes the CNN nearly exact.} The 3D-conv CNN\arch{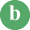} introduces the front end that the structured models below share (Appendix Figure~\ref{fig:dsconv}). A $3{\times}3$ convolution maps each cell to an 18-dimensional situation code, and a $2{\times}3{\times}3$ convolution over two adjacent frames maps each observed transition to a 36-dimensional (situation, outcome) code. A linear aggregator sums the transition codes into a tally (18 situations $\times$ 2 outcomes), and a gated head looks up the current situation in it. With analytic detectors the model is exact at L3; learned, it reaches 99.7\%. Every inexact learned run misses the detector for one of the two rarest situations (each about 0.2\% of cells), and more budget, width, or batch does not repair it (Appendix~\ref{app:mech}).

\subsection{Transformer: look-back and look-ahead pairing}
\label{sec:attn}

\begin{figure}[t]
\centering
\includegraphics[width=\linewidth]{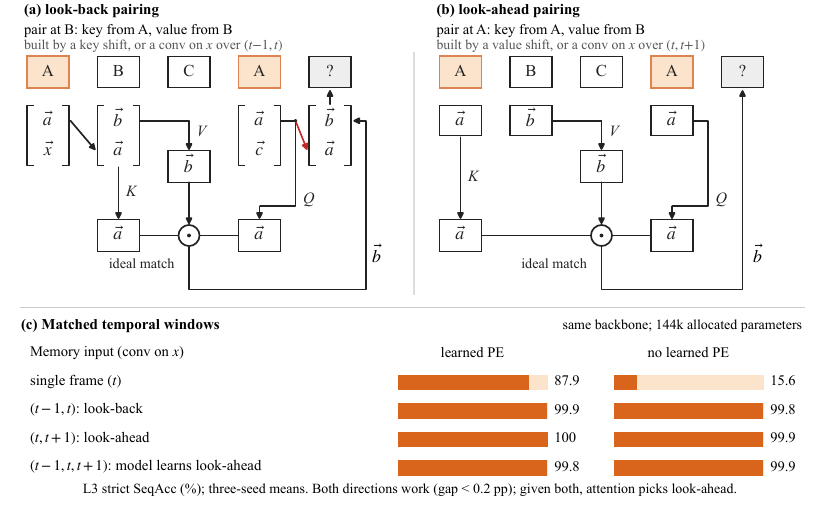}
\caption{\textbf{Two kinds of induction.} (a)~Look-back pairing stores the observed pair A$\to$B at B, whose key carries its predecessor A. (b)~Look-ahead pairing stores it at A, whose value carries its successor B. Either way, the second A retrieves B. (c)~One memory-attention backbone (144k parameters, identical initialization) whose memory is a convolution on the input $x$ over one of four temporal windows, with and without learned positions (L3 SeqAcc, \%). A KV shift builds (b) by index; a convolution can build either. Every window that spans a transition works, in either direction; the six symmetric-window models (two position settings, three seeds) attend in the look-ahead convention.}
\label{fig:twoinductions}
\end{figure}

The classic induction circuit of language models pairs by looking back \citep{elhage2021mathematical, olsson2022incontext}. In the sequence A B C A (Figure~\ref{fig:twoinductions}a), a previous-token head copies A into the residual stream at B, so the key at B carries A and its value carries B; the second A matches that key and retrieves B. In look-ahead pairing (Figure~\ref{fig:twoinductions}b), each token keeps its own content as key and its value carries its successor, so the pair is stored at A. Both are causal here: the current frame only queries, and every stored pair belongs to a frame whose successor has been observed. A short convolution builds either pair \citep{so2021primer, fu2023hungry, arora2023zoology, allenzhu2025canon}, and so does a KV shift \citep{xu2024kvshift}.

\textbf{Supplying the pair matters more than scale.} The 3D-conv transformer\arch{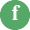} turns each observed transition into one token whose key is the situation and whose value is the outcome, and retrieves with a single head at a frozen temperature of 20 and no positional encoding. With 3.8k learned parameters it is exact at L1--L3 on every seed. The KV-shift transformer\arch{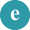} builds the same pair by index---key from frame $t$, value from frame $t{+}1$ \citep{lin2019tsm, peng2023rwkv}---and is exact on every seed with 2.6k parameters, over two thousand times fewer than the 6.6M-parameter standard transformer, which is not exact (Figure~\ref{fig:main}). The 2D-conv transformer\arch{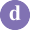} must learn the pair: it encodes each frame separately and passes the unpaired tokens, with learned positions, through two attention layers. With 142k parameters it reaches 88.8\% (Table~\ref{tab:e12pos}, Appendix~\ref{app:e12}).

\textbf{The 2D-conv transformer learns to pair by looking back, but its retrieval stays soft.} On every seed its first hop attends to the same cell in the previous frame and almost never to the next one (Table~\ref{tab:e12circuit}, Appendix~\ref{app:e12}). Its learned temperature settles below 1, so attention stays spread and small first-step errors compound over the rollout. Errors concentrate on situations observed only once or twice, where many near matches outweigh a lone exact one \citep{singh2024needs, velickovic2024softmax}. On an unfiltered stream, freezing the temperature, five times the budget, and doubled width do not repair it (Appendix~\ref{app:mech}).

\textbf{With the backbone fixed, any window that spans a transition works} (Figure~\ref{fig:twoinductions}c). One memory-attention model, with 144{,}050 allocated parameters and identical initialization in every arm, builds its memory from one of four temporal windows, with or without learned positions. Every window spanning a transition---$(t{-}1,t)$, $(t,t{+}1)$, or $(t{-}1,t,t{+}1)$---reaches at least 99.8\% with or without positions, against 87.9\% and 15.6\% for the single frame $(t)$ (Figure~\ref{fig:e18pair}, Appendix~\ref{app:e12}). The two directions perform alike, and the six symmetric-window models ($(t{-}1,t,t{+}1)$; two position settings, three seeds) attend mostly to the historical situation itself rather than to its successor (Table~\ref{tab:e18anchors}, Appendix~\ref{app:e12}): they match A and read what followed it, the look-ahead convention of the KV-shift transformer.

\textbf{The same binding rescues the standard transformer, given positions.} Attaching to each raster token its cell's $3{\times}3$ neighbourhood in the previous frame (704 parameters at d64) lifts the d64 transformer from 25.8\% to 99.9\%, and d128 from 85.6\% to 99.9\% (the ``last-frame $3{\times}3$'' rows of Figure~\ref{fig:main}). With no position encoding at all, the same decoders reach only 8.1--8.2\% (Table~\ref{tab:e15pair}, Appendix~\ref{app:e12}). This feature supplies spatial alignment and the temporal pair at once; the matched windows above isolate the temporal part. Binding is not a drop-in replacement everywhere: as the front end of the structured diffusion model of Section~\ref{sec:stability}, the KV shift reaches 49.0\% against 97.7\% for the convolution (Appendix~\ref{app:mech}). What carries across designs is the requirement---each observed outcome bound to a situation the query can match---not one implementation.

\subsection{Level 4: when induction needs a prior}
\label{sec:silent}

\textbf{When the evidence does not contain the answer, the model needs a prior and a route to apply it.} At L4 the queried situation never appears in the observed frames; the answer follows from the partner row's observed outcome only through knowledge of the rule family---here, that the two rows are negations of each other. Matching alone cannot supply it. A tally over all evidence is one route: the 3D-conv CNN reaches 98.8\%, and the structured diffusion model (Section~\ref{sec:stability}), same tally, 93.1\%. A single attention read is not: the KV-shift and 3D-conv transformers, exact at L3, drop to 42.2\% and 71.5\%. Four stacked reads are another route: 99.5\% for the 3D-conv transformer under the ordinary loss, and 99.5\% for the KV-shift transformer once the loss is balanced over the 0.18\% of training cells that need the partner (Figure~\ref{fig:e21}). The standard d128 decoder with the previous-frame neighbourhood reaches 97.2\%. Targeted data matters more than training time: at the natural rate of the event, three times the steps never match the 10\% dose (Figure~\ref{fig:silent}, Appendix~\ref{app:mech}; cf.\ \citealp{chan2022data, reddy2024mechanistic}). A matched inductive bias lowers how much targeted data a prior needs, a cost that matters most where such data are scarce.

\section{Temporal stability: settle each frame before it is used}
\label{sec:stability}

\begin{figure}[t]
\centering
\begin{minipage}[t]{0.68\linewidth}
\centering
\includegraphics[width=\linewidth]{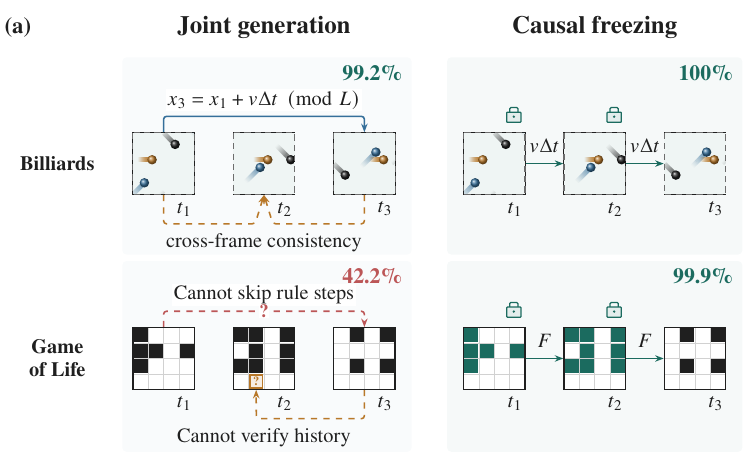}
\end{minipage}\hfill
\begin{minipage}[t]{0.30\linewidth}
\centering
\includegraphics[width=\linewidth]{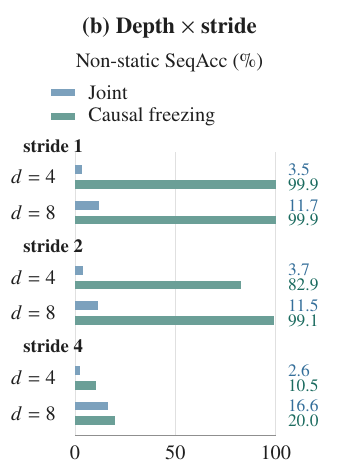}
\end{minipage}
\caption{\textbf{Billiards and the Game of Life under the same denoiser architecture.} (a)~One plain denoiser trained per world, weights fixed across samplers, sampled jointly or with causal freezing at eight calls per rollout (SeqAcc). Billiards can be extrapolated across frames; in Life one call cannot skip rule steps, and a later frame cannot verify an earlier one. The drawings illustrate these properties, not model outputs. (b)~Life, both samplers, across network depth $d$ and the number of rule steps per displayed frame (stride).}
\label{fig:billiards}
\end{figure}

Rollout adds a third requirement: a predicted frame that becomes the input for the next must be settled first. Diffusion models make this testable, because the same weights can generate all frames jointly or one at a time. We use two settings: structured denoisers on unseen rules, and a plain denoiser on the Game of Life and on billiards.

\textbf{The standard diffusion model is correct for two or three frames and then collapses.} It completes no held-out rollout on any seed, and only 24.9\% even on the Game of Life. Under uniform joint denoising no predicted frame is ever settled, so each prediction is revised together with the frames it depends on. Joint denoising could carry out the computation in principle: given an exact transition $F$, the recurrence $\hat{x}_k^{(r+1)}=F(\hat{x}_{k-1}^{(r)})$ with $\hat{x}_0^{(r)}=x_0$ extends a correct prefix by one frame per round. Training does not find it.

\textbf{With the weights fixed, changing only the order of denoising lifts held-out completion from 4.7\% to 76.3\%.} We train two structured denoisers (main and curriculum recipes) with independent per-frame noise levels, a backward-looking canvas encoder, and a clean anchor frame (Appendix Figure~\ref{fig:dsdiff}). Sampled jointly over 49 network calls, they complete 4.7\% of held-out rollouts. Sampled with \emph{causal freezing}\arch{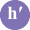}---denoise each frame in order, then keep it as clean context \citep{chen2024diffusion}---the same weights reach 76.3\% with 48 calls, and 88.2\% with one call per frame. Retrained under the common recipe of Figure~\ref{fig:main}, the same swap lifts the denoiser from 4.5\% to 27.9\%: the direction holds under both recipes, the size does not. The weights already encode the transition; the joint schedule does not turn it into a continuation.

\textbf{Freezing matters for the Game of Life, not for billiards} (Figure~\ref{fig:billiards}a). We train one plain depth-8 3D-convolutional denoiser (196k parameters) separately on free billiards and on the Game of Life, and change only the sampler, at eight network calls either way. Joint generation reaches 99.2\% on billiards against 100\% with freezing, but 42.2\% on Life against 99.9\%. A billiard ball's later position follows from its current position and velocity in one step. Life must be composed one local update at a time, and a later Life state can have several predecessors, so it cannot uniquely confirm an earlier guess.

\textbf{Depth does not rescue joint generation; freezing degrades only when one call must compose several rule steps} (Figure~\ref{fig:billiards}b). Going from four to eight layers leaves joint generation far behind freezing. When each displayed frame is two rule steps after the last, depth-8 freezing still reaches 99.1\%; at four steps, both samplers fail on most worlds. The successful sampler behaves as one-step autoregression trained with a denoising objective: a shallow denoiser composes few rule steps per call, and freezing supplies the intermediate states that joint training does not learn to keep.

\textbf{Training matched to the schedule gives 99.4\%.} A clean prefix, one noisy target frame, noise beyond it, and loss on that frame only---the diffusion analogue of exposure-bias remedies \citep{bengio2015scheduled, huang2025selfforcing}---define the structured diffusion model\arch{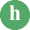}: 99.4\% on held-out rules with 30k parameters (Appendix~\ref{app:timeline}). Per-frame noise levels and ordered generation already appear in streaming video diffusion \citep{chen2024diffusion, ruhe2024rolling, kim2024fifo, yin2024causvid, song2025history, magi2025}; our experiments show their role in exact continuation.

\textbf{Left to choose, the denoisers commit frames in causal order.} Committing the most confident frame at each stage \citep{chang2022maskgit, kim2025train} selects exactly the order $1{\to}8$ in about 98\% of worlds and matches the imposed order's accuracy; least-confident-first and random orders stay below 5\% (Table~\ref{tab:e12order}, Appendix~\ref{app:e12}). These weights carry directional priors (the backward encoder, the anchor), so this is order selection under those priors, not discovery of causality. Order need not mean one frame per call: the plain denoiser commits several Life frames per call and keeps 99.8\% (Appendix~\ref{app:billiards}).

\section{What the three properties have in common}
\label{sec:common}

Each property is one link in Figure~\ref{fig:infoflow}: spatial locality brings a cell's neighbours together (step~1), temporal locality (momentum induction) pairs each observed situation with its outcome (steps~2--3), and temporal stability settles each prediction before later ones use it (the rollout loop). Reachability can be checked before training; learnability only by training, and that is where every standard model within our budget fell short. When the evidence does not contain the answer (Section~\ref{sec:silent}), the missing link is a prior about the world's rule family, and the same two tests apply: does a route exist, and does training use it?

This suggests a short checklist for a new world. (i)~Write down which cells and frames must meet, and which parts of the future depend on which. (ii)~Check that some part of the model can see both members of each pair. (iii)~Supply the links that training is unlikely to find: a position scheme, a pairing, a commitment order. It is not a new method, and is close in spirit to algorithmic alignment \citep{xu2020reason} and to Canon layers \citep{allenzhu2025canon}. Settling intermediate states resembles the role of chain of thought in transformers \citep{feng2023cot, merrill2024cot}.

\section{Future Work}
\label{sec:outlook}

Our evidence comes chiefly from a deterministic family of binary automata on an $8\times8$ grid; the physical and video directions below remain tests of transfer. Small exploratory experiments on continuous dynamics (Appendix~\ref{app:bridge}, Figures~\ref{fig:bridgetasks}--\ref{fig:bridgebars}) already suggest that the answer depends on the dynamics: with shared weights, causal freezing helps Burgers but not linear advection--diffusion, while the locality and pairing results are mixed.

\textbf{Exact physical generation.} We will test the three properties in richer systems---interacting particles and continuous dynamics---where a known simulator makes fidelity checkable over a specified horizon: exact agreement for discrete states, declared tolerances for continuous ones. This separates executing the dynamics from the visual plausibility of a rendered rollout; the experiments of Appendix~\ref{app:bridge} are a first step.

\textbf{From latent dynamics to rendered video.} Instead of showing CA states directly to the model, a rendered automaton keeps the simple rules while adding textures, lighting, and camera motion. A video model could encode the observed frames, infer transitions with the two-frame pairing in a learned latent space, and decode the predicted states into frames; we will ask whether this improves state-level consistency without sacrificing visual quality, relative to an otherwise matched predictor (Appendix~\ref{app:outlook}). The question is whether the same information flow survives perception and rendering.

\section*{Ethics Statement}
This work studies small synthetic cellular-automaton systems generated by a
documented simulator. It involves no human subjects, personal data, or
deployed systems, and we foresee no direct ethical risk from the models or the
data.

\section*{Reproducibility Statement}
Every number in this paper is regenerated from per-run logs: each registered
run records its command, seeds, hardware, a SHA-256 fingerprint of the exact
training code, and the checksums of the pinned evaluation corpora used, and
every reported number is transcribed from these logs without manual editing.
One supplementary probe---the attention-depth ladder at L4---was registered from a
retrospectively preserved source snapshot rather than a prelaunch code
fingerprint; its predictions and checkpoints passed the same audits.
Training data are generated deterministically from (seed, index) by a
version-frozen simulator, so no training corpus needs to be shipped;
evaluation corpora are pinned files with SHA-256 sidecars that regenerate
identically on any machine. Protocol, architecture, and recipe details are in
Appendices~\ref{app:protocol}--\ref{app:recipes}. The full
codebase---simulator, single-entry training script, evaluation and analysis
tooling, and the pinned evaluation corpora---has been staged locally for
release. Public availability and a stable archive link will be confirmed
in the arXiv version after the release checks are complete.

\section*{Statement on the Use of AI Assistance}
In this work, we used generative AI tools (large-language-model assistants)
for: providing feedback on research methodology and experiment design, with
several of the paper's techniques---including the constructor-swap and
sampler-swap controls---originating in human--AI collaborative design
sessions; helping develop the conceptual framework (the information-flow
requirements and the audit protocol); implementing methods (experiment scripts,
model code, and analysis tooling); interpreting results (drafting diagnostic
summaries from run logs); assisting with translation between the authors'
working language and English; and drafting and editing parts of the paper for
readability. Additionally, we used them for identifying and verifying
literature, formatting references, creating figure-generation scripts, and
suggesting experimental parameters. We did not use generative AI tools to
generate synthetic datasets---all data are produced by a deterministic,
version-frozen simulator from (seed, index)---and this work contains no
mathematical proofs, so proof-related assistance is not applicable. We have
reviewed all AI-assisted work: every experimental number was checked by the
authors against the registered run logs; citation keys were checked against
the bibliography, and bibliographic metadata was checked against available
arXiv, CrossRef, and publisher records where applicable; AI-assisted code is
covered by a test suite, including bit-identity tests for every compatibility
flag; and the authors directed and adjudicated all design proposals. We take
full responsibility for the final content of this work, including text,
claims, and artifacts produced with the aid of generative AI.


\appendix

\section{Related work}
\label{app:related}

\paragraph{Cellular automata and rule learning.}
Cellular automata make global dynamics from local rules
\citep{wolfram1983statistical, gardner1970fantastic}. Convolutional networks
can implement cellular-automaton updates by construction
\citep{gilpin2019cellular}, although learning even the Game of Life from
examples can be difficult \citep{springer2021hard}. Neural cellular automata
learn a parameterized local update \citep{mordvintsev2020growing}; related
sequence-model studies include shortcuts on finite automata
\citep{liu2023shortcuts}, elementary one-dimensional rules with
rule-disjoint free runs \citep{burtsev2024learning}, recursive prediction
under the fixed Game of Life rule \citep{berkovich2025lifegpt}, and
two-dimensional rule-supplied forecasting alongside inverse rule inference
\citep{berkovich2026automatagpt}. Our setting isolates prefix-conditioned
continuation on held-out rules: queried situations are covered by observed
transitions, and success requires every cell in every predicted frame to be
correct (\S\ref{sec:task}). We diagnose known devices under this exact metric
without claiming priority for learning across rules.

\paragraph{Spatial locality and trainability.}
Convolutions expose neighbouring cells directly, whereas a raster sequence
requires the model to recover two-dimensional adjacency, including boundary
wrap-around. Rotary position embeddings originated in sequence models
\citep{su2024roformer}; two-dimensional variants are established for image
tokens \citep{heo2024rotary}. The choice of position scheme exemplifies how
architecture and target computation should align \citep{battaglia2018relational,
xu2020reason, velickovic2020neural}; input parameterization can change which
algorithm a transformer learns \citep{mcleish2024arithmetic}. In our
controlled spatial comparisons, the same architecture can represent the rule
but its reliability changes when positions or local inputs expose the grid
structure (\S\ref{sec:spatial}). Thus a locality benefit in this testbed
does not imply a general inability of attention to compute local updates.

\paragraph{Induction and context-dependent retrieval.}
Induction-head accounts describe a previous-token operation followed by
matching and copying \citep{elhage2021mathematical, olsson2022incontext};
controlled work studies how such behaviour forms and depends on data
\citep{singh2024needs, edelman2024statistical, chan2022data,
reddy2024mechanistic}. Theory characterizes representational and training
conditions for induction on restricted tasks \citep{sanford2024onelayer,
ekbote2025one, nichani2024transformers}. Our coverage-filtered
cellular-automaton task calls for matching a local situation in the observed
frames and retrieving its successor.
The attention patterns we report are consistent with this two-step account,
but observational attention and approximate matches do not identify a unique
causal circuit (\S\ref{sec:attn}, Appendix~\ref{app:e12}).

\paragraph{Putting situations beside outcomes.}
Temporal shifts in video and sequence models \citep{lin2019tsm,
peng2023rwkv}, short convolutions for recall \citep{fu2023hungry,
arora2023zoology}, and shifted keys and values for induction
\citep{xu2024kvshift} provide precedents for placing information from
adjacent times in one representation. Our convolution over adjacent frames,
index-aligned key--value shift, and previous-frame neighbourhood input adapt
such existing devices to observed situation--successor pairs. They differ in
where the pair is stored, but each makes the transition available before
retrieval. Matched temporal-window controls test the pairing requirement
without assigning special status to one direction (\S\ref{sec:temporal}).
The observed gain concerns learnability under this task and training budget,
not a new pairing operation.

\paragraph{Diffusion and commitment order.}
Denoising diffusion underlies video generation
\citep{ho2020denoising, ho2022video}; per-frame noise levels and ordered or
streaming generation already appear in Diffusion Forcing, rolling diffusion,
FIFO-Diffusion, and causal video generation \citep{chen2024diffusion,
ruhe2024rolling, kim2024fifo, yin2024causvid}. Decoding order also matters
in masked discrete generation \citep{kim2025train}. In our controlled
comparison, one denoiser's weights are held fixed while joint denoising and
causal freezing use different schedules with 49 and 48 network calls,
respectively (\S\ref{sec:stability}). The result tests whether settling each
predicted frame before later frames use it helps exact continuation in these
worlds; it neither introduces frame ordering nor establishes a universal
preference for it.

\paragraph{World models and exact evaluation.}
Othello-GPT probes a learned board-state representation
\citep{li2023emergent}. Other work on learned world models asks whether
plausible trajectories reflect the underlying dynamics \citep{ha2018world, vafa2024evaluating,
vafa2025foundation}; video and physical-reasoning evaluations likewise probe
the gap between appearance and physical prediction
\citep{bear2021physion, bansal2024videophy, motamed2025physicsiq}.
Measurement choices can change apparent capability
\citep{schaeffer2023emergent}, while video quality measures such as FVD
\citep{unterthiner2018towards} answer a different question from whether
every predicted cell in every frame follows the rule. We therefore report
strict sequence accuracy on rule-disjoint, coverage-filtered continuations
and check that the observed evidence determines their targets
(\S\ref{sec:task}). The local pairing and frame-order results suggest
hypotheses for broader generators, rather than verified transfer beyond this
controlled family (\S\ref{sec:outlook}).

\section{Protocol and reporting}
\label{app:protocol}
\label{app:reporting}

This appendix fixes the rule split, the test corpora, and the reporting rules.

\paragraph{Rule indexing and split.}
A rule is an 18-bit integer whose bit $9s+n$ is the next state of situation $(s,n)$. The split is a fixed-seed permutation of the $262{,}144$ indices, because table entries are index bits and a parity or modulo split would tie some entries to one half.

\paragraph{Coverage and the self-consistency filter.}
Without the filter, 33\% of L3 trajectories query a situation that the observed frames never show. If $u$ counts such situations, a uniform prior on the unseen rule entries gives the realized continuation posterior mass $2^{-u}$, with mean 0.8099. This is not an upper bound on SeqAcc: averaged over the 2048 observed prefixes, the most probable continuation carries mass 0.8159, or 0.8732 under the prior restricted to the held-out half---conditional references, not measured accuracies. The analytic 3D-conv CNN (Section~\ref{sec:conv}) scores 0.8032 on the unfiltered corpus and 1.0000 on its covered subset. Rejection sampling keeps 67\% of draws at L2 and L3. Masking the loss on unobserved situations instead of filtering the stream is neutral ($\Delta = 0.0001$, paired).

\paragraph{Level 4 corpora.}
Level 4 worlds come from held-out rules with the prior of Section~\ref{sec:task} built in: the two situations with opposite centre cells and $n\in\{3,4\}$ black neighbours have opposite outcomes, $r(1,n)=1-r(0,n)$. L4B, the Level 4 test of the main text, holds 768 trajectories; in each, a queried situation never appears among the observed transitions while its centre-flipped partner does. They were selected from 3.86M draws, so the event's natural rate is about $2\times10^{-4}$. Level 4 training therefore replaces 10\% of each batch (the \emph{10\% dose}) with such trajectories, drawn from a pool of 65{,}536 training-half trajectories generated with a seed of its own; \emph{natural rate} denotes training without the dose. L4A, a companion test of 2048 trajectories, instead fixes four rule rows across the family and queries one of them without showing it, so the model must recall the family's fixed value.

\paragraph{Pixel-level versus sequence-level accuracy.}
At 0.999 per pixel, independent errors would still spoil about two in five eight-frame rollouts on the $8\times8$ torus ($1-0.999^{512}\approx0.4$), and the metric cannot separate a model that has the rule from one that is approximately right.

\paragraph{Reporting conventions.}
Numbers are final-step values on exponential-moving-average (EMA) weights, without checkpoint selection, over three preregistered seeds (42/43/44). Main-text numbers are three-seed means in percent, truncated, not rounded, to one decimal: 0.9995 shows as 99.9, and only an exact 1.0 shows as 100. Appendix tables give the per-seed values behind them. Figure cells with fewer registered seeds are marked where they appear. Two choices used validation data, never test data: the checkpoint of one continuous-dynamics experiment (Appendix~\ref{app:bridge}) and the number of attention reads in Figure~\ref{fig:e21}. Data are generated deterministically from (seed, index); evaluation corpora carry SHA-256 sidecars and hold 512 trajectories at L1, 2048 at L2 and L3, and 768 at L4 (L4B). Models train on the filtered stream with bf16 autocast and are evaluated in fp32 unless a table says otherwise; runs on the unfiltered stream appear only where labelled.

\subsection{Preregistered predictions and corrections}
\label{app:process}

Two predictions registered for the standard transformer failed on the filtered stream (Table~\ref{tab:scladder}). The first was that it would stay near zero on held-out rules ($\approx$0--5\%); d128 reaches 0.9263/0.7646/0.8784 on L3. The second was that accuracy would rise monotonically with width; d128 is the best rung, while d192 and d256 each leave one seed at the marginal predictor. Its other half, that no rung would be exact, held.

\section{Models and training}
\label{app:models}

\begin{figure}[t]
\centering
\includegraphics[width=\linewidth]{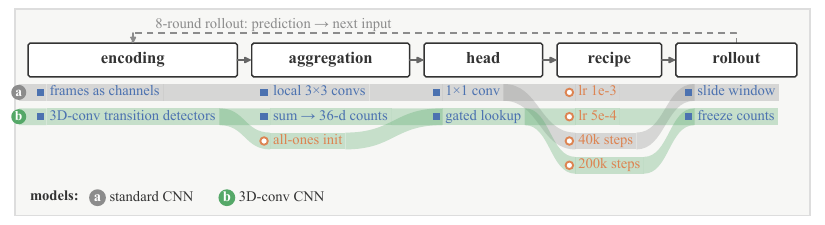}
\caption{\textbf{Design space of the convolutional models.} Columns are pipeline stages; each coloured band traces one model. Squares are architectural choices, circles training choices.}
\label{fig:dsconv}
\end{figure}

\begin{figure}[t]
\centering
\includegraphics[width=\linewidth]{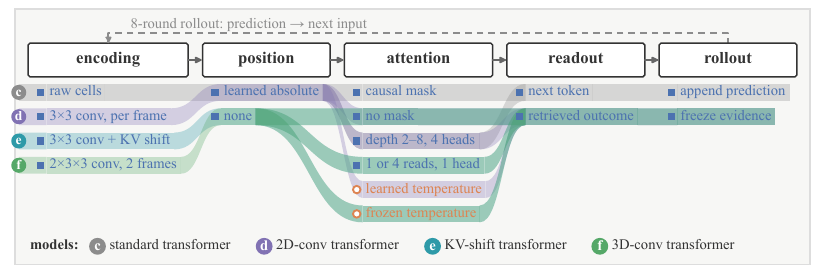}
\caption{\textbf{Design space of the attention models}, read as Figure~\ref{fig:dsconv}. The encoding column names each model: raw cells, a per-frame $3{\times}3$ convolution, the same with a KV shift, or a $2{\times}3{\times}3$ convolution over two adjacent frames.}
\label{fig:dsattn}
\end{figure}

\begin{figure}[t]
\centering
\includegraphics[width=\linewidth]{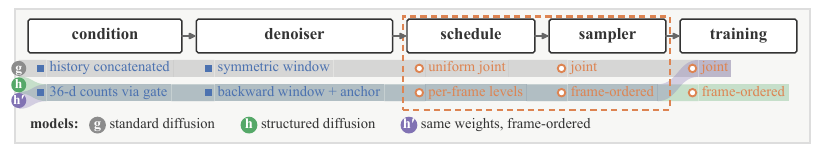}
\caption{\textbf{Design space of the diffusion models}, read as Figure~\ref{fig:dsconv}. Circles are training and sampling choices; frame-ordered sampling is causal freezing.}
\label{fig:dsdiff}
\end{figure}

Figures~\ref{fig:dsconv}, \ref{fig:dsattn}, and~\ref{fig:dsdiff} place every model in its family's design space; each band carries the model's letter from Figure~\ref{fig:main}.

\paragraph{Detection front end.}
The three structured families share two detectors. A $3{\times}3$ circular convolution with two channels, followed by two small ReLU layers, maps each cell to an 18-dimensional situation code (one-hot at the analytic solution); the same stack with a time kernel of 2 and three channels maps each adjacent frame pair to a 36-dimensional (situation, outcome) code per cell. Two ReLU layers are needed because ``exactly $n$ black neighbours'' is not linearly separable from a single convolution.

\paragraph{How each family uses the codes.}
The 3D-conv CNN gates each transition channel, sums it with a grouped spatial-sum pyramid, combines the 7 frame pairs linearly per channel, and applies a fixed $\log(1+x)$; a per-cell head reads the resulting tally beside the situation code. The 3D-conv transformer makes each of the 448 observed transitions a token (key: its situation channels; value: its outcome) and queries with the current situation code. Scores are $20\,q\cdot k$, leaking about $10^{-6}$ of the softmax weight onto non-matching one-hot tokens; a head MLP ($36\to24\to1$) reads the retrieved value and the code. The structured diffusion model noises each predicted frame by a variance-preserving first-order autoregressive chain (retention 0.9, $T=50$), with targets $\pm5$ and belief $\hat{x}_0=5\tanh(u/2)$ for network output $u$. Its canvas encoder looks only backward (frame $k$ reads frames $k{-}1$ and $k$), the last observed frame is a clean anchor, and a bilinear gate combines canvas features with the tally. Training matches causal freezing: clean ground truth before the denoised frame, that frame at its noise level, noise beyond, and loss on that frame only.

\paragraph{Parameter counts.}
3D-conv CNN 17{,}994 (analytic 15{,}361); 3D-conv transformer 3{,}798 (analytic 913); KV-shift transformer 2{,}625; 2D-conv transformer 141{,}698 (104{,}834 without learned positions); structured diffusion model 29{,}963; standard transformer 165{,}825, 924{,}801, 2{,}866{,}753, and 6{,}581{,}505 at d64, d128, d192, and d256 (2, 4, 6, and 8 layers), plus 704 at d64 for the last-frame $3{\times}3$ input; standard CNN 116{,}097; standard diffusion model 85{,}697 (depth 4); plain depth-8 denoiser of the billiards comparison 196{,}417.

\paragraph{Supervision.}
All models train on their ordinary loss and are given structure and initialization only, with one exception: the evidence-balanced loss of Figure~\ref{fig:e21} (the L4 cells marked * in Figure~\ref{fig:main}) averages the loss separately over direct-evidence and partner-only cells, a training-side task label that never enters the model as input. The filtered stream and the 10\% L4 dose choose training trajectories and add no loss term.

\paragraph{Scaffolds.}
Given, not learned: the 3D convolution (time kernel 2); in the 3D-conv transformer, the frozen attention temperature 20 and no positional encoding; in the learned 3D-conv CNN and structured diffusion model, an analytic all-ones initialization of the aggregation subnets; in the structured diffusion model, the backward receptive field, the clean anchor frame, and the per-frame noise schedule. Diffusion has no analytic version; its results rest on learned weights.

\paragraph{Training recipe.}
Adam ($\beta_2=0.999$), gradient clipping at 1.0, 500 warmup steps then cosine decay to 10\%, and EMA 0.999; all evaluation uses the EMA weights. All transformers, the standard CNN, and the standard diffusion model train for 40k steps at batch 512 and learning rate $10^{-3}$; the 3D-conv CNN trains for 200k steps at batch 128 and the structured diffusion model for 200k steps at batch 2048, both at learning rate $5\times10^{-4}$. Recipe variants and per-cell exceptions are in Appendix~\ref{app:recipes}.

\section{Per-seed results and training budgets}
\label{app:recipes}

\begin{figure}[t]
\centering
\includegraphics[width=\linewidth]{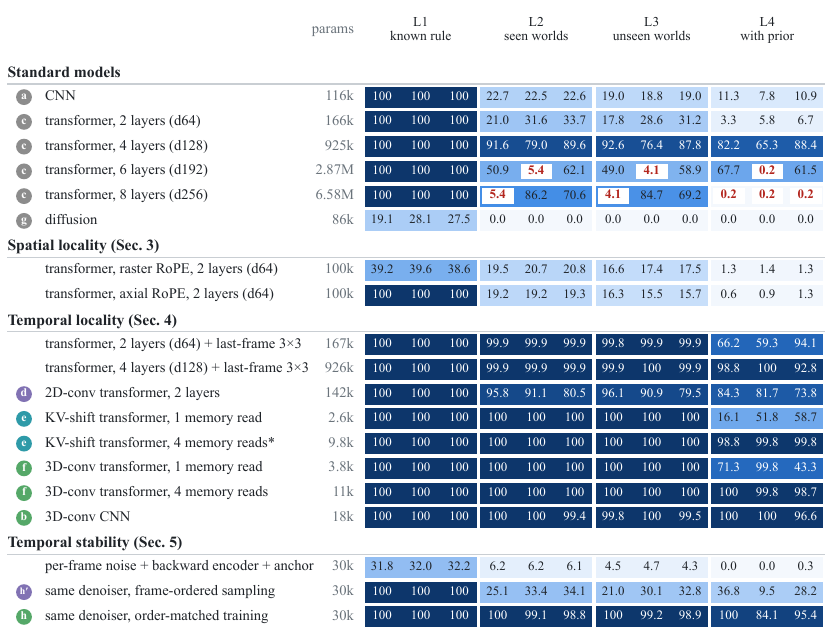}
\caption{\textbf{Main results, per seed.} The rows and columns of Figure~\ref{fig:main}; each cell gives SeqAcc (\%) for seeds 42, 43 and 44 at the final step, truncated to one decimal, and is shaded by their mean. Red: standard-transformer runs that never left the marginal predictor. The 200k-step runs of the standard models are in Tables~\ref{tab:recipes} and~\ref{tab:scladder}.}
\label{fig:mainseeds}
\end{figure}

Figure~\ref{fig:mainseeds} gives the three seeds behind every cell of Figure~\ref{fig:main}, in the same layout. Table~\ref{tab:main} repeats the L1, L3 and L4 values at four decimals with exact parameter counts and lists the protocol exceptions; the four-read rows are detailed in Figure~\ref{fig:e21} and the sampler comparison of the structured denoiser in Appendix~\ref{app:timeline}. Table~\ref{tab:twometrics} gives the per-seed values of Figure~\ref{fig:twometrics}, and Table~\ref{tab:recipes} the controls on training budget and recipe.

\begin{table}[t]
\centering
\caption{\textbf{Main results by number.} Strict SeqAcc (fraction) at the final step on EMA weights. Three numbers are seeds 42/43/44 and a single number is the single run of an analytic model; for the d64, d192 and d256 standard transformers the entry is the min--max range over seeds 42/43/44 (per-seed L3 values in Table~\ref{tab:scladder}, all per-seed values in Figure~\ref{fig:mainseeds}). Training uses the filtered stream, bf16 and the reported recipes of Appendix~\ref{app:models}; L4 is scored on the 768 forced test trajectories after training with the 10\% dose (Appendix~\ref{app:protocol}). Exceptions: the L4 cells of the 3D-conv CNN, the 3D-conv transformer and the structured diffusion model are fp32 runs, the 3D-conv CNN's at learning rate $10^{-3}$; the 3D-conv transformer's L4 cells are the model without the tally (with the tally: 0.9974\,/\,0.9987\,/\,0.9180, Figure~\ref{fig:silent}); L3 values of the 2D-conv, KV-shift and 3D-conv transformers trained on the unfiltered stream in fp32 are in Appendix~\ref{app:mech}. ``---'' marks cells not registered.}
\label{tab:main}
\small
\adjustbox{max width=\columnwidth}{%
\begin{tabular}{lrccc}
\toprule
Model & Params & L1 & L3 & L4 \\
\midrule
3D-conv CNN (analytic) & 15,361 & 1.0 & 1.0000 & --- \\
3D-conv CNN (learned) & 17,994 & 1.0\,/\,1.0\,/\,1.0 & 0.9980\,/\,1.0000\,/\,0.9951 & 1.0000\,/\,1.0000\,/\,0.9661 \\
3D-conv transformer (analytic) & 913 & 1.0 & 1.0000 & --- \\
3D-conv transformer (learned) & 3,798 & 1.0\,/\,1.0\,/\,1.0 & 1.0\,/\,1.0\,/\,1.0 & 0.7135\,/\,0.9987\,/\,0.4336 \\
Structured diffusion model & 29,963 & 1.0\,/\,1.0\,/\,1.0 & 1.0000\,/\,0.9927\,/\,0.9893 & 1.0000\,/\,0.8411\,/\,0.9544 \\
\midrule
2D-conv transformer & 141,698 & 1.0\,/\,1.0\,/\,1.0 & 0.9614\,/\,0.9097\,/\,0.7954 & 0.8438\,/\,0.8177\,/\,0.7383 \\
KV-shift transformer & 2,625 & 1.0\,/\,1.0\,/\,1.0 & 1.0000\,/\,1.0000\,/\,1.0000 & 0.1615\,/\,0.5182\,/\,0.5872 \\
\midrule
Standard transformer d128 & 924,801 & 1.0\,/\,1.0\,/\,1.0 & 0.9263\,/\,0.7646\,/\,0.8784 & 0.8229\,/\,0.6536\,/\,0.8841 \\
Standard transformer d64 $+$ last-frame $3{\times}3$ & 166,529 & 1.0\,/\,1.0\,/\,1.0 & 0.9985\,/\,0.9995\,/\,0.9995 & 0.6628\,/\,0.5938\,/\,0.9414 \\
Standard transformer d128 $+$ last-frame $3{\times}3$ & 926,209 & 1.0\,/\,1.0\,/\,1.0 & 0.9990\,/\,1.0000\,/\,0.9995 & 0.9883\,/\,1.0000\,/\,0.9284 \\
Standard transformer d64 & 165,825 & 1.0\,/\,1.0\,/\,1.0 & 0.1787--0.3120 & 0.0339--0.0677 \\
Standard transformer d192 & 2,866,753 & 1.0\,/\,1.0\,/\,1.0 & 0.0410--0.5894 & 0.0026--0.6771 \\
Standard transformer d256 & 6,581,505 & 1.0\,/\,1.0\,/\,1.0 & 0.0410--0.8477 & 0.0026 $\times$3 \\
Standard CNN & 116,097 & 1.0\,/\,1.0\,/\,1.0 & 0.1909\,/\,0.1880\,/\,0.1904 & 0.1133\,/\,0.0781\,/\,0.1094 \\
Standard diffusion model & 85,697 & 0.1914\,/\,0.2812\,/\,0.2754 & 0.0000\,/\,0.0000\,/\,0.0000 & 0.0000\,/\,0.0000\,/\,0.0000 \\
\bottomrule
\end{tabular}}
\end{table}

\begin{table}[t]
\centering
\small
\caption{\textbf{Per-seed values of Figure~\ref{fig:twometrics}.} L3, fractions to four decimals (a value below 1 is never shown as 1.0000), seeds 42/43/44. Literature-architecture baselines adapted to our data and rollout protocol: NCA \citep{mordvintsev2020growing}, the one-dimensional transformer of \citet{burtsev2024learning}, AutomataGPT \citep{berkovich2026automatagpt}, LifeGPT \citep{berkovich2025lifegpt}, and Diff-NCA \citep{kalkhof2024frequency}. Pixel-level accuracy is teacher-forced (one step ahead) except in the two diffusion rows marked $\dagger$, which score the rollout; SeqAcc always scores the full rollout.}
\label{tab:twometrics}
\adjustbox{max width=\columnwidth}{%
\begin{tabular}{lll}
\toprule
Model & Pixel-level accuracy & SeqAcc \\
\midrule
NCA (growing) & 0.7715\,/\,0.7711\,/\,0.7718 & 0.1387\,/\,0.1021\,/\,0.1377 \\
Standard CNN & 0.9625\,/\,0.9645\,/\,0.9636 & 0.1909\,/\,0.1880\,/\,0.1904 \\
3D-conv CNN & 0.9999\,/\,1.0000\,/\,0.9999 & 0.9980\,/\,1.0000\,/\,0.9951 \\
Burtsev 1-D transformer & 0.8243\,/\,0.9815\,/\,0.9768 & 0.0386\,/\,0.0405\,/\,0.0405 \\
AutomataGPT & 0.9303\,/\,0.9416\,/\,0.9351 & 0.1528\,/\,0.1689\,/\,0.1655 \\
LifeGPT ($4\times128$) & 0.9999\,/\,0.9982\,/\,0.9980 & 0.9497\,/\,0.5098\,/\,0.4683 \\
KV-shift transformer & 1.0000\,/\,1.0000\,/\,1.0000 & 1.0000\,/\,1.0000\,/\,1.0000 \\
Diff-NCA & 0.7932\,/\,0.7888\,/\,0.7913 & 0.1470\,/\,0.1489\,/\,0.1484 \\
Standard diffusion model$^\dagger$ & 0.6167\,/\,0.6218\,/\,0.6211 & 0.0000\,/\,0.0000\,/\,0.0000 \\
Structured diffusion model$^\dagger$ & 1.0000\,/\,0.9992\,/\,0.9984 & 1.0000\,/\,0.9927\,/\,0.9893 \\
\bottomrule
\end{tabular}}
\end{table}

\paragraph{LifeGPT configuration.}
The LifeGPT row of Figure~\ref{fig:twometrics} \citep{berkovich2025lifegpt} is the $4\times128$ configuration (1,052,800 parameters), designated in advance and within the 0.5--1.5M budget for prior-work configurations. It keeps raster next-token generation, the original split-half partial RoPE, and forgetful causal masking during training only. It uses our binary vocabulary, the common training protocol of 40k steps at batch 512, and greedy evaluation of the final EMA weights. This is architecture-faithful retraining under our protocol. Across three seeds its L3 SeqAcc is $64.26\pm26.68\%$ (sample SD).

\paragraph{Training budgets and recipe choices.}
Table~\ref{tab:recipes} lists the controls on training budget, learning rate and precision. At five times the steps, the standard CNN ends at 0.2070/0.2065/0.2051: its floor is converged, not budget-starved.

Both the 3D-conv CNN and the structured diffusion model were first registered at learning rate $10^{-3}$; the reported $5\times10^{-4}$ replaced it after one seed of each ended low, and both recipes are listed with all their seeds. For the CNN, the 0.8350 cell is a late loss spike, which the lower rate removes. For the diffusion model, the registered rate leaves one of five seeds stuck at 0.9175; neither continued annealing nor a warm restart from that checkpoint frees it, and only retraining from initialization at $5\times10^{-4}$ lifts it to 0.9805. The remaining gap on that seed survives batch 4096 and changes of width, consistent with the rare-situation tail of Appendix~\ref{app:mech}. In fp32 and in bf16 alike, both models stay inexact on one or two seeds; the choice of precision does not remove the tail.

\begin{table}[t]
\centering
\caption{\textbf{Training budgets and recipe choices.} Strict L3 SeqAcc (fraction) at the final step on EMA weights, one value per listed seed. The second column lists every difference from the model's row in Table~\ref{tab:main}; indented rows modify the nearest unindented row above them.}
\label{tab:recipes}
\small
\adjustbox{max width=\columnwidth}{%
\begin{tabular}{llll}
\toprule
Model & Difference from Table~\ref{tab:main} & Seeds & L3 \\
\midrule
Standard CNN & 200k steps ($5\times$), fp32 & 42/43/44 & 0.2070\,/\,0.2065\,/\,0.2051 \\
\midrule
3D-conv CNN & learning rate $10^{-3}$, fp32 & 42/43/44 & 0.9937\,/\,0.8350\,/\,0.9946 \\
3D-conv CNN & fp32 & 42/43/44 & 0.9917\,/\,1.0000\,/\,0.9814 \\
\midrule
Structured diffusion model & learning rate $10^{-3}$, fp32 & 42/43/44/45/46 & 0.9990\,/\,1.0\,/\,0.9175\,/\,0.9941\,/\,1.0 \\
\quad from the stuck checkpoint & continued annealing & 44 & 0.9165 \\
\quad from the stuck checkpoint & warm restart & 44 & 0.9229 \\
Structured diffusion model & fp32 & 42/43/44 & 1.0000\,/\,1.0000\,/\,0.9805 \\
\quad same recipe & batch 4096 & 44 & 0.9873 \\
\bottomrule
\end{tabular}}
\end{table}

\section{Spatial locality in the raster transformer}
\label{app:serialized}

The standard transformer of Section~\ref{sec:task} reads a world's sixteen frames as one frame-major raster sequence of 1024 tokens with learned absolute positions, trains with a causal next-token loss under teacher forcing, and rolls out greedily; depth grows with width, from d64 with 2 layers to d256 with 8 (Appendix~\ref{app:models}). Runs use 40k steps and seeds 42/43/44 unless stated.

\subsection{Rotary positions on the Game of Life}

Table~\ref{tab:e14} gives the d64 transformer on the Game of Life (L1) under six position encodings, adapting one-dimensional RoPE \citep{su2024roformer} and two-dimensional visual RoPE \citep{heo2024rotary} to the toroidal grid. With rotary positions along the raster sequence (1D RoPE) it completes 39.1\% of rollouts, and 95.4\% of its teacher-forced errors fall on the 28 border cells (43.75\% of the board). In raster order a cell's left neighbour in the previous frame is 65 tokens back, but in column 0 it wraps to the end of the row and is 57 tokens back. Both offsets always land on some token, so an encoding that sees only raster offsets cannot tell which applies to which cell; absolute positions, or rotation by row and column separately (2D axial RoPE), can. This concerns what the encodings can express, not a mechanism located in the trained models.

Every two-dimensional variant repairs the failure (99.4--100\%), including the period-16 and half-integer detuned controls, whose periods are deliberately wrong. The result supports two-dimensional position, not the period, as what matters; the preregistered prediction that the board's own period would beat the other two-dimensional variants was not met. All arms keep the raster causal mask, and time is never periodic. The learned reference has more parameters than the rotary arms, so it is a reference, not a matched control. The rule is known, so rule inference is not tested.

\begin{table}[t]
\centering
\includegraphics[width=\linewidth]{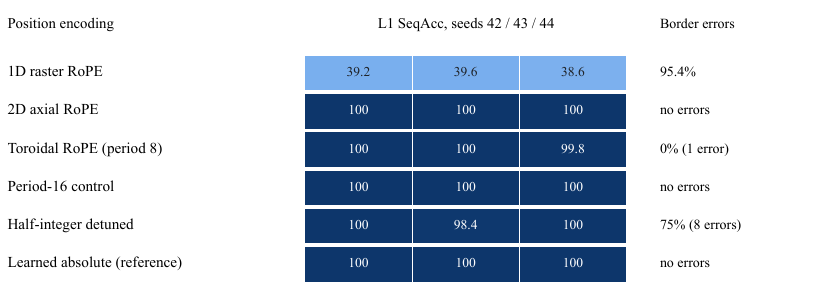}
\caption{\textbf{Rotary positions on the Game of Life (L1).} Standard transformer d64, 2 layers. Each SeqAcc cell prints seeds 42/43/44 in percent, truncated to one decimal; its three colour blocks encode those seeds separately on a common 0--100\% scale. Border error share is a separate teacher-forced diagnostic. Five rotary variants have identical parameter counts (100{,}289) and initialization; the learned reference has 165{,}825 parameters.}
\label{tab:e14}
\end{table}

\subsection{Rotary positions on unseen rules}

\begin{figure}[t]
\centering
\includegraphics[width=\linewidth]{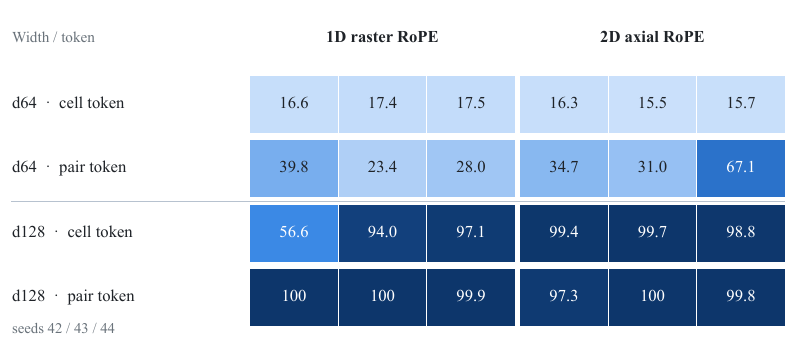}
\caption{\textbf{Rotary positions and previous-frame neighbourhoods on unseen rules (L3).} Standard transformer L3 SeqAcc (\%): each cell prints seeds 42/43/44, truncated to one decimal, in three separately shaded blocks on a common 0--100\% scale. A cell token carries only its own cell; a pair token adds the previous-frame $3{\times}3$ neighbourhood. With pair tokens and 1D RoPE at d64, 98.2--99.5\% of teacher-forced errors fall on the 28 border cells. Full two-decimal values remain in the E22 results archive.}
\label{fig:e22}
\end{figure}

Figure~\ref{fig:e22} crosses the two rotary schemes with two kinds of token on unseen rules (L3), at d64 and d128. A pair token also carries the cell's previous-frame $3{\times}3$ neighbourhood (the binding of Section~\ref{sec:attn}), so pair arms have slightly more parameters. With cell tokens, 2D RoPE lifts d128 from 82.6\% to 99.3\%; at d64 neither scheme exceeds 18\%.

Four contrasts were preregistered; each is met only if two of three seeds clear its per-seed threshold. (i)~2D minus 1D RoPE, pair tokens: $+13.9$ pp at d64, $-0.9$ pp at d128; not met. (ii)~Pair minus cell tokens, 2D RoPE: $+28.4$ pp at d64, where only one seed clears, and $-0.3$ pp at d128; not met. (iii)~2D minus 1D RoPE, cell tokens, registered as descriptive: $-1.3$ pp at d64, $+16.7$ pp at d128, higher on every d128 seed. (iv)~Border share of teacher-forced errors, pair tokens with 1D RoPE: met at d64 (mean 98.95\%); undefined at d128, where two seeds make no teacher-forced errors. The experiment thus does not establish two-dimensional rotary positions as a robust improvement on unseen rules; the d128 cell-token gain stays descriptive. At d64 the rotary pair arms reach only 30.4\% and 44.3\%, against 99.9\% with learned positions (Table~\ref{tab:e15pair}).

\subsection{Width ladder}
\label{app:scladder}

\begin{figure}[t]
\centering
\includegraphics[width=0.88\columnwidth]{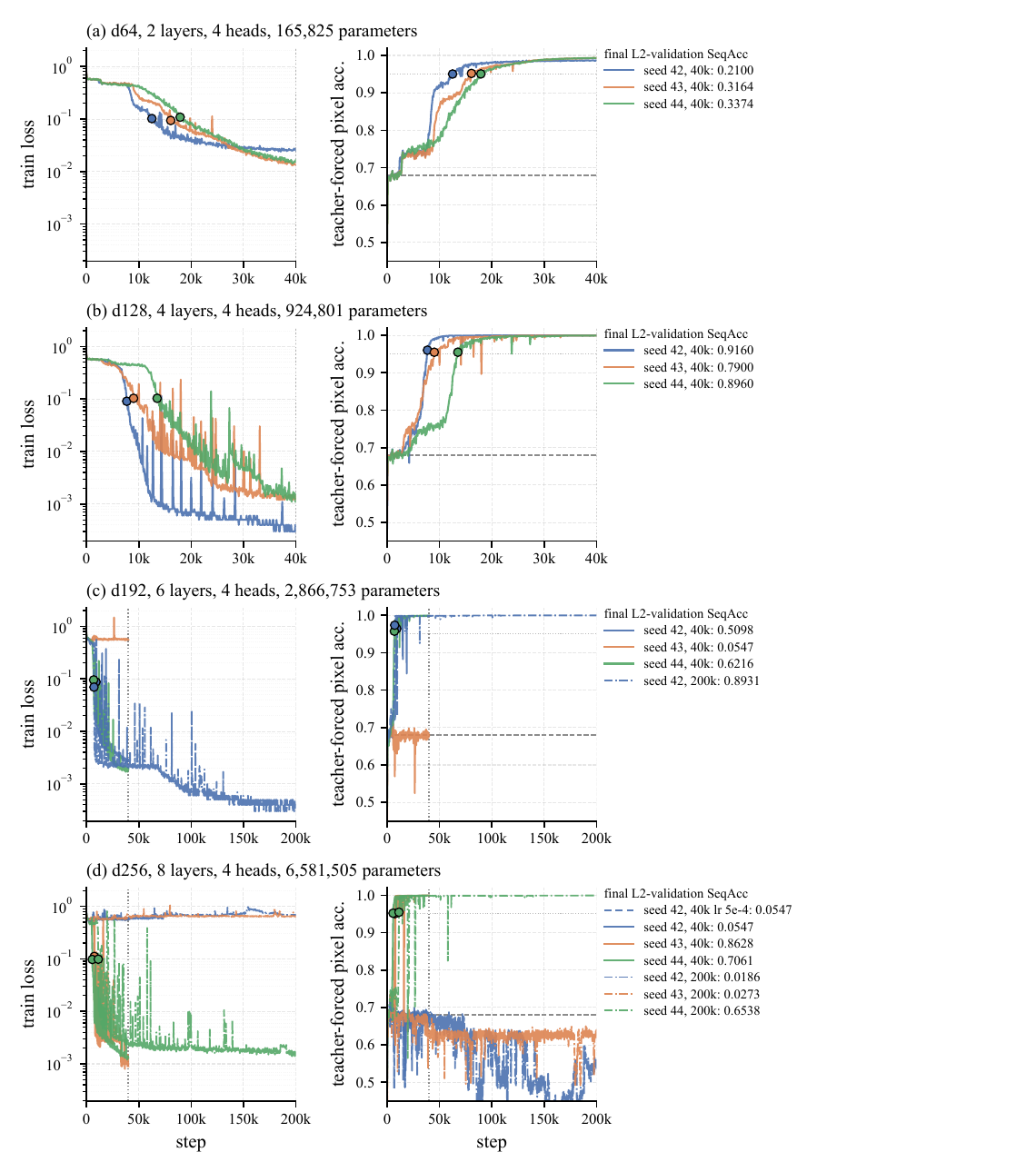}
\caption{\textbf{Training curves of the width ladder.} Training loss and teacher-forced pixel accuracy, one row per width; colours are seeds 42/43/44; solid lines are 40k-step runs, dash-dotted lines 200k-step runs, and the dashed line the halved learning rate. Circles mark escape, the first step with teacher-forced pixel accuracy $\geq 0.95$; stuck runs are flat from ${\sim}1$k steps.}
\label{fig:scladder}
\end{figure}

Table~\ref{tab:scladder} gives the standard transformer with learned positions at four widths (filtered stream, learning rate $10^{-3}$). A run either leaves the marginal predictor, which ignores the evidence and scores 0.0410 on L3, before 18k steps or never (Figure~\ref{fig:scladder}); no run escapes between 40k and 200k steps. At 40k steps d192 and d256 each leave one seed of three at the floor. In separate 200k-step runs (middle block) d192 seed 42 improves (L3 0.8828 against 0.4902 at 40k), while d256 seeds 42 and 43 stay stuck, although seed 43 escaped in its 40k run, so the seed alone does not decide escape. Halving the learning rate to $5\times10^{-4}$ leaves d256 seed 42 at the floor too (L3 0.0405 against 0.0410; not tabulated).

Accuracy does not rise monotonically with width: d128 is the best rung on average (85.6\% on L3), and no arm is exact. L2 validation (seen rules) tracks L3 in every run, so escaped runs infer the rule rather than recall it. The bottom block, a single-seed depth-by-width check, is descriptive: six layers instead of two raise d64 from 0.2100 to 0.6616; at d192, two layers give 0.4614 against 0.5098.

\begin{table}[t]
\centering
\small
\caption{\textbf{Width ladder of the standard transformer.} SeqAcc (fraction of rollouts exact) on L2 validation and L3. Top: 40k steps, seeds 42\,/\,43\,/\,44. Middle: separate 200k-step runs. Bottom: depth by width at seed 42, L2 validation only.}
\label{tab:scladder}
\adjustbox{max width=\columnwidth}{%
\begin{tabular}{llll}
\toprule
Arm & Steps & L2 validation & L3 \\
\midrule
d64, 2 layers  & 40k & 0.2100 / 0.3164 / 0.3374 & 0.1787 / 0.2861 / 0.3120 \\
d128, 4 layers & 40k & 0.9160 / 0.7900 / 0.8960 & 0.9263 / 0.7646 / 0.8784 \\
d192, 6 layers & 40k & 0.5098 / 0.0547 / 0.6216 & 0.4902 / 0.0410 / 0.5894 \\
d256, 8 layers & 40k & 0.0547 / 0.8628 / 0.7061 & 0.0410 / 0.8477 / 0.6929 \\
\midrule
d192, 6 layers, seed 42 & 200k & 0.8931 & 0.8828 \\
d256, 8 layers, seed 42 & 200k & 0.0186 & 0.0151 \\
d256, 8 layers, seed 43 & 200k & 0.0273 & 0.0234 \\
d256, 8 layers, seed 44 & 200k & 0.6538 & 0.6372 \\
\midrule
d64, 2 layers, seed 42  & 40k & 0.2100 & --- \\
d64, 6 layers, seed 42  & 40k & 0.6616 & --- \\
d192, 2 layers, seed 42 & 40k & 0.4614 & --- \\
d192, 6 layers, seed 42 & 40k & 0.5098 & --- \\
\bottomrule
\end{tabular}}
\end{table}

\section{Diagnostics for temporal locality, Level 4, and temporal stability}
\label{app:mech}
\label{app:e12}

All measurements below use the final-step EMA weights of registered runs, evaluated in fp32 on the self-consistent test corpora of Section~\ref{sec:task}. Attention and order probes change no weights; the position-free arms and the matched-window experiment are fresh training. Protocols and primary probes were frozen before launch; later diagnostics are labelled post hoc.

\subsection{Convolution: the rare-situation tail}

Every learned 3D-conv CNN run below 1.0 (Table~\ref{tab:main} and Appendix~\ref{app:recipes}) concentrates its first errors on one of the two rarest situations (each about 0.2\% of cells), for which no detector forms. Which run misses a detector depends on the seed: more budget, width, or batch does not repair it, and the learning rate helps only weakly.

\subsection{Which pairing the 2D-conv transformer learns}

We capture attention per block and head at the teacher-forced first prediction step on all 2048 L3 worlds (Table~\ref{tab:e12circuit}). Unmasked attention lets the first hop point either way; on every seed it points back, to the same cell in the previous frame, as if each pair were stored at its outcome (look-back pairing). The final match is content-correct under that reading and near chance under the look-ahead one. The pattern does not identify a unique mechanism. What stays soft is retrieval: errors concentrate where the queried situation is rare in the observed frames, and wrong cells spread their attention over situations one live neighbour away. The evidence-count rows were added after the first pass and are descriptive.

\begin{table}[h]
\centering
\caption{\textbf{Attention readout of the 2D-conv transformer.} Filtered training stream; teacher-forced first prediction step on all 2048 L3 worlds; seeds 42/43/44, whose L3 SeqAcc is 0.9614/0.9097/0.7954. Same-cell entries give the largest same-cell attention mass over heads; the uniform level is 0.002.}
\label{tab:e12circuit}
\small
\adjustbox{max width=\columnwidth}{%
\begin{tabular}{lccc}
\toprule
Measure & s42 & s43 & s44 \\
\midrule
Learned temperature (3D-conv transformer: frozen at 20) & 0.846 & 0.750 & 0.662 \\
Same-cell attention to the previous frame (look-back) & 0.472 & 0.445 & 0.388 \\
Same-cell attention to the next frame (look-ahead; uniform 0.002) & 0.003 & 0.002 & 0.003 \\
Final match content-correct, look-back reading & 0.9995 & 0.9986 & 0.9955 \\
Final match content-correct, look-ahead reading & 0.1836 & 0.1734 & 0.2109 \\
Top-1 attention mass & 0.138 & 0.144 & 0.176 \\
First-step cell accuracy & 0.99979 & 0.9995 & 0.99832 \\
\midrule
Error rate, situation seen once among the observed transitions (276 cells) & 0.069 & 0.109 & 0.228 \\
Error rate, seen twice (400 cells) & 0.020 & 0.058 & 0.150 \\
Error rate, seen 3--4 times (983 cells) & 0 & 0.011 & 0.052 \\
Error rate, seen 17 or more times (119{,}722 cells) & 0 & 0 & 0 \\
Mean exact matches among the observed transitions, wrong / correct cells & 1.3 / 61.0 & 1.8 / 61.0 & 3.1 / 61.1 \\
Attention to one-neighbour-off situations, wrong $\div$ correct & 24.5 & 37.7 & 15.7 \\
\bottomrule
\end{tabular}}
\end{table}

Soft retrieval resists the obvious fixes. On the unfiltered stream the seed-42 model completes 0.8047 of L3 rollouts; freezing the temperature at 20 (0.793, and 0.758 after five times the budget) and doubling the width with $3.4\times$ the parameters (0.778) do not repair it. Each is a single run.

\subsection{Positions and pairing across designs}

Table~\ref{tab:e12pos} sets the 2D-conv transformer beside the two constructors that are given the pair. Removing its token and query position tables (recipe otherwise identical) makes its retrieval a function of the set of observed tokens, so the first hop cannot find the previous frame. It then ends below the standard CNN (L3 0.1553/0.1558/0.1538; L2 0.1851/0.1841/0.1851), and the readout of Table~\ref{tab:e12circuit} shows why: attention to the previous and to the next frame both stay at the uniform level (0.002 on every seed), top-1 mass is 0.02, and first-step cell accuracy is 0.73.

\begin{table}[h]
\centering
\caption{\textbf{Pairing across designs.} L3 SeqAcc, filtered stream, seeds 42/43/44. Architectures, temperatures, and parameter counts differ; index alignment is not a learned positional embedding.}
\label{tab:e12pos}
\small
\adjustbox{max width=\columnwidth}{%
\begin{tabular}{lrlc}
\toprule
Model & Params & Pairing and position structure & L3 SeqAcc \\
\midrule
KV-shift transformer & 2{,}625 & index-aligned pairs; no learned positions & 1.0 / 1.0 / 1.0 \\
3D-conv transformer & 3{,}798 & adjacent-frame pairs; no learned positions & 1.0 / 1.0 / 1.0 \\
2D-conv transformer & 141{,}698 & unpaired input; learned positions & 0.9614 / 0.9097 / 0.7954 \\
2D-conv transformer & 104{,}834 & unpaired input; no learned positions & 0.1553 / 0.1558 / 0.1538 \\
\bottomrule
\end{tabular}}
\end{table}

Constructors given the pair need no learned positions, but pairing does not make positions dispensable in every design (Table~\ref{tab:e15pair}). Without positions, the standard d128 transformer (793{,}729 parameters) has only its causal mask as an order signal. Causal decoders can recover position from the mask \citep{haviv2022nope, kazemnejad2023nope}, but within 40k steps none of the three seeds leaves the marginal predictor (L3 0.0410 and L2 0.0547 on every seed; teacher-forced pixel accuracy 0.66--0.67), and the previous-frame neighbourhood on each token does not rescue either width.

\begin{table}[h]
\centering
\caption{\textbf{Previous-frame neighbourhood and learned positions in the standard transformer.} L3 SeqAcc (\%), mean over seeds 42/43/44. Removing the position table also changes parameter count and initialization; the causal mask and the neighbourhood input still carry temporal and spatial structure. All cells share the recipe and were scored in one evaluation pass: the learned-position d64 cells and the d128 own-cell cell are the checkpoints of Table~\ref{tab:main} re-scored here (d128 own cell 0.9282/0.7681/0.8906, against 0.9263/0.7646/0.8784 in Table~\ref{tab:main} and 85.6\% in the main text), and the d128 cell with the previous-frame neighbourhood and learned positions comes from three reference runs of the same recipe (1.0/1.0/0.9990).}
\label{tab:e15pair}
\small
\adjustbox{max width=\columnwidth}{%
\begin{tabular}{llrr}
\toprule
Width & Token input & Learned positions & No position encoding \\
\midrule
d64 & own cell & 25.90 & 4.10 \\
d64 & $+$ previous-frame $3{\times}3$ & 99.92 & 8.12 \\
d128 & own cell & 86.23 & 4.10 \\
d128 & $+$ previous-frame $3{\times}3$ & 99.97 & 8.27 \\
\bottomrule
\end{tabular}}
\end{table}

\subsection{Matched temporal windows}

One memory-attention model isolates the temporal window (Figure~\ref{fig:twoinductions}c). It keeps the 2D-conv transformer's backbone ($d=64$, two unmasked four-head memory-attention blocks, a learned temperature initialized to 1) and builds its memory with a $3{\times}3{\times}3$ convolution whose input gates expose one of four windows at each of the eight observed anchors. Unavailable temporal neighbours are zero, never wrapped. Every arm allocates 144{,}050 parameters with identical same-seed initialization, but unused slices and gated position tables leave different weights in use: equal allocation is not equal effective capacity. All 24 runs (four windows, two position settings, seeds 42/43/44) train on the filtered stream for 40k updates and are scored on all 2048 unseen worlds; within-seed comparisons share hardware.

\begin{figure}[t]
\centering
\includegraphics[width=\linewidth]{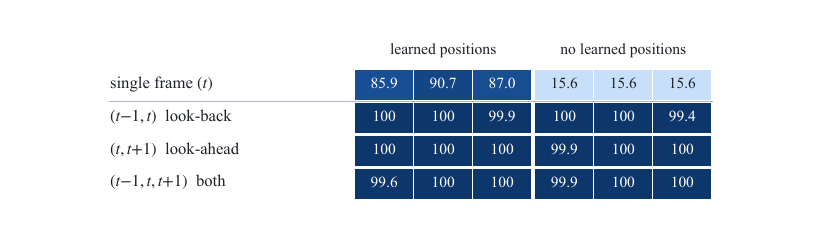}
\caption{\textbf{Matched temporal windows with and without learned positions.} L3 SeqAcc (\%) at the final step. Each cell shows seeds 42/43/44, truncated to one decimal, in three separately shaded blocks on the same 0--100\% scale. The top row uses one frame; the remaining windows span an observed transition. All arms allocate 144{,}050 parameters with identical same-seed initialization; the weights in use differ. Full-precision values and sample SD are registered in \texttt{docs/RESULTS.md}.}
\label{fig:e18pair}
\end{figure}

The preregistered performance predictions are met (Figure~\ref{fig:e18pair}): the forward window $(t,t{+}1)$ without positions exceeds 95\% with an on/off gap below two points, and it improves on the single frame $(t)$ by 12.061 points with learned positions and 84.310 without. The direction prediction is inconclusive: forward minus backward is 0.033 and 0.163 points, inside the preregistered two-point region.

Table~\ref{tab:e18anchors} asks where the most-attended memory token sits relative to the matching historical situation A: at A itself, whose successor supplies the outcome (cause-side, the look-ahead convention), or one frame later, at A's outcome (effect-side, the look-back convention). The six symmetric-window models (two position settings, three seeds each) all favour the cause side. A post-hoc partition into cause-only, effect-only, both, and neither keeps this reading: in every symmetric arm the effect-only fraction stays below 0.9\% (maximum 0.86594\%). The single frame with learned positions shows the opposite, predominantly effect-side pattern (0.974/0.995/0.913), like the look-back pairing of Table~\ref{tab:e12circuit}.

These are anchor associations, not an identified key/value mechanism, and the convolution window is not a literal KV shift. A final token has passed through two unmasked blocks, which can relocate evidence: the forward window with learned positions completes every rollout at seed 42 with a cause-side match of only 0.323 (0.999 at seeds 43 and 44), and the backward window at seed 43 is exact with an effect-side match of 0.077 and 50.6\% of final attention on boundary anchors. High match rates also need not mean large attention mass.

\begin{table}[h]
\centering
\small
\caption{\textbf{Which historical anchor matches the query?} Fraction of first-step queries whose most-attended interior memory token sits at the matching situation (cause-side) or one frame after it, at its outcome (effect-side). All 2048 L3 worlds, seeds 42/43/44, common interior anchors 1--6. The two readings overlap and are observational: two memory-attention blocks can relocate evidence before the final readout.}
\label{tab:e18anchors}
\adjustbox{max width=\columnwidth}{%
\begin{tabular}{llcc}
\toprule
Window & Learned positions & Cause-side top-1 & Effect-side top-1 \\
\midrule
$(t)$ & on & 0.256 / 0.249 / 0.239 & 0.974 / 0.995 / 0.913 \\
$(t)$ & off & 0.090 / 0.136 / 0.110 & 0.091 / 0.125 / 0.086 \\
$(t,t{+}1)$ & on & 0.323 / 0.999 / 0.999 & 0.163 / 0.242 / 0.229 \\
$(t,t{+}1)$ & off & 0.563 / 0.999 / 0.999 & 0.101 / 0.107 / 0.106 \\
$(t{-}1,t)$ & on & 0.242 / 0.137 / 0.251 & 0.999 / 0.077 / 0.970 \\
$(t{-}1,t)$ & off & 0.211 / 0.234 / 0.158 & 0.999 / 0.659 / 0.725 \\
$(t{-}1,t,t{+}1)$ & on & 0.931 / 0.817 / 0.999 & 0.211 / 0.203 / 0.207 \\
$(t{-}1,t,t{+}1)$ & off & 0.806 / 0.904 / 0.999 & 0.064 / 0.096 / 0.159 \\
\bottomrule
\end{tabular}}
\end{table}

\subsection{Binding across designs and training streams}

The pairing requirement carries across designs; a particular constructor need not. Replacing the $2{\times}3{\times}3$ convolution of the structured diffusion model by the index-aligned KV shift, with the 200k-step recipe otherwise unchanged, gives L3 SeqAcc 0.5962/0.2490/0.6250 (mean 49.0\%) against 1.0000/1.0000/0.9316 (97.7\%) for the convolution retrained on the same machine. The preregistered prediction that the two constructors are interchangeable was not met. Parameter counts (29{,}206 against 29{,}963) and initialization trajectories differ, so this is a recipe-specific comparison of constructors, not an isolated cause.

The training stream separates the attention constructors too. On the unfiltered stream about a third of training trajectories query a situation the observed frames never show, whose label is a fair coin. L3 SeqAcc at seeds 42/43/44, filtered stream (bf16 training) against unfiltered stream (fp32 training): 2D-conv transformer 0.9614/0.9097/0.7954 against 0.8047/0.8853/0.8926; KV-shift transformer 1.0000/1.0000/1.0000 against 0.8442/1.0000/0.9297; 3D-conv transformer 1.0000 on every seed of both. The index-shift constructor alone is exact on one stream and not the other; since precision also differs, the comparison is descriptive.

\subsection{Level 4: evidence withheld}

\begin{figure}[t]
\centering
\includegraphics[width=\linewidth]{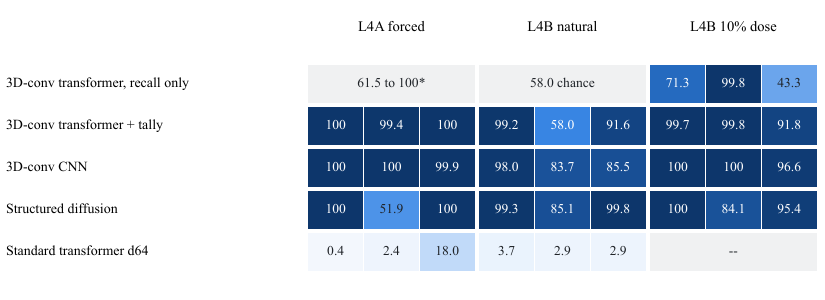}
\caption{\textbf{Level 4 grids: evidence withheld.} SeqAcc (\%) for seeds 42/43/44, separately shaded within each cell. L4A holds 2048 forced arbitrations; L4B holds 768 forced pair inferences. Natural rate and 10\% dose denote distinct training streams. Grey cells are single-run or missing: * recall-only L4A is 61.52\% at 40k steps and 100\% at 120k or with head width 48; its natural-rate L4B score is 58.07\%, approximately chance. Standard d64 has no dose arm; its natural-rate values come from separate runs evaluated on the unforced L4B corpus. Full precision remains in the results registry.}
\label{fig:silent}
\end{figure}

Figure~\ref{fig:silent} gives the full grids behind Section~\ref{sec:silent}. L4B is the level called L4 in the main text, with the centre-flip prior of Section~\ref{sec:task} (Appendix~\ref{app:protocol}). L4A fixes four rule rows across the whole family and queries one of them without showing it, so the answer is the family's fixed value. ``Recall only'' is the 3D-conv transformer of the main text; ``$+$ tally'' adds the 3D-conv CNN's tally to its head.

Training time does not substitute for the dose. At the natural rate the tally model trains three times as long as with the dose (120k against 40k steps) and still ends below its dose runs on every seed; single seed-42 natural-rate runs of the 3D-conv CNN and the structured diffusion model at 600k steps, three times their 200k budget, also stay below their dose cells. The 3D-conv CNN's dose cells are higher than its natural-rate cells on every seed, while the structured diffusion model, trained for 200k steps either way, scores about the same with and without the dose.

\begin{figure}[t]
\centering
\includegraphics[width=\linewidth]{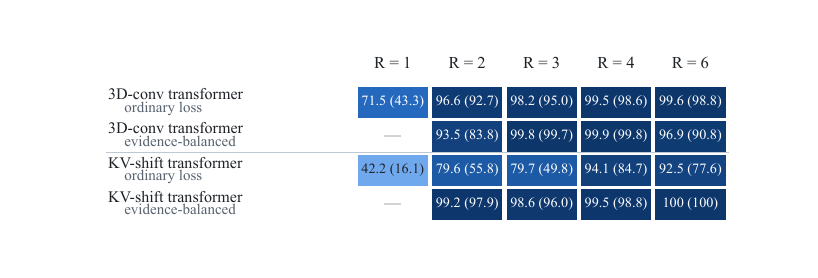}
\caption{\textbf{Attention depth at L4.} Strict L4B SeqAcc (\%) on 768 held-out worlds needing the partner inference. Each cell prints the three-seed mean (worst seed), truncated to one decimal, and is shaded by its mean. $R$ sequential attention reads over frozen evidence memory; 40k steps and 10\% dose throughout. Four reads were chosen on training-half validation before this test. Evidence-balanced loss averages direct-evidence and partner-only cells separately, using a training-side label. $R=1$ values are the single-read models of Table~\ref{tab:main}; exact per-seed values and parameter counts remain in the results registry.}
\label{fig:e21}
\end{figure}

Figure~\ref{fig:e21} adds attention depth. Extra reads close the L4 gap for the 3D-conv transformer under the ordinary loss, and for the KV-shift transformer only with the evidence-balanced loss. Depth and parameter count grow together, so the ladder does not separate them.

\subsection{Structured diffusion: sampler swap and matched training}
\label{app:timeline}

The two structured denoisers of Section~\ref{sec:stability} (main and curriculum recipes) are trained for 1M steps with independent per-frame noise levels, a backward-looking canvas encoder, and a clean anchor frame (Appendix Figure~\ref{fig:dsdiff}); under joint sampling they stayed at 0.046--0.049 throughout training. Table~\ref{tab:samplerswap} reads their final-step weights under four samplers.

\begin{table}[h]
\centering
\caption{\textbf{Same weights, four samplers.} L3 SeqAcc of the final-step weights of the two structured denoisers trained with independent per-frame noise levels (seed 42 each). Causal freezing denoises each frame in turn and keeps it as clean context; its call budget is split over the eight frames.}
\label{tab:samplerswap}
\small
\adjustbox{max width=\columnwidth}{%
\begin{tabular}{llcc}
\toprule
Sampler & Network calls & Main & Curriculum \\
\midrule
Joint denoising & 49 & 0.0483 & 0.0469 \\
Causal freezing & 48 & 0.7402 & 0.7866 \\
Causal freezing, one call per frame & 8 & 0.8750 & 0.8892 \\
Causal freezing, full chain per frame & 49 per frame & 0.7622 & 0.8081 \\
\bottomrule
\end{tabular}}
\end{table}

The weights already encode the transition: given the true frames before it, a single predicted frame is extended with pixel accuracy 0.9986/0.9991, and 0.93--0.96 of frames are perfect, with no decline along the rollout. If frames failed independently, rollouts would complete at the product of these rates, 0.63/0.71; causal freezing does better, so the errors cluster within worlds.

Matched training trains on the inputs that causal freezing produces: clean ground truth before the frame being denoised, that frame at its noise level, noise beyond it, and loss on that frame only. The resulting structured diffusion model (29{,}963 parameters) reaches L3 1.0000/1.0000/0.9805 in fp32 and 1.0000/0.9927/0.9893 in the bf16 re-run of Table~\ref{tab:main}, and 1.0 at L1 on all three seeds.

\subsection{Commitment order}

At each of eight stages, one denoiser call scores every uncommitted frame by its mean confidence $|\tanh(u/2)|$; the most confident frame then runs its full chain and is committed. This is the confidence ordering of masked generative models \citep{chang2022maskgit, kim2025train}, applied to frames without a prescribed order (Table~\ref{tab:e12order}). The per-frame-level weights, never trained for an order, choose exactly the causal order in 98.1\%/98.7\% of worlds and match the imposed order's accuracy; the commonest exception (10 worlds per checkpoint) commits predicted frames 2, 4, 6, 8 before 1, 3, 5, 7. These checkpoints already have a backward-looking canvas encoder, paired evidence, and a clean last-observed anchor, so this is order selection under those priors, not discovery of causality. The imposed-order column is a separate sampling pass of the full-chain sampler; its gap to Table~\ref{tab:samplerswap} is sampling noise.

\begin{table}[h]
\centering
\caption{\textbf{Confidence-ordered commitment.} L3 SeqAcc; ``Causal'' is the share of worlds whose chosen order is exactly $1{\to}8$. Per-frame-level rows: the two denoisers of Table~\ref{tab:samplerswap}; matched-training rows: the structured diffusion model at seeds 42/43/44.}
\label{tab:e12order}
\small
\adjustbox{max width=\columnwidth}{%
\begin{tabular}{lccccc}
\toprule
Weights & Most confident first & Causal & Imposed $1{\to}8$ & Least confident first & Random \\
\midrule
Per-frame levels, main & 0.7739 & 0.981 & 0.7749 & 0.0366 & 0.0386 \\
Per-frame levels, curriculum & 0.8169 & 0.987 & 0.8169 & 0.0386 & 0.0400 \\
Matched training, s42 & 1.0000 & 1.000 & 1.0000 & 0.0000 & 0.0000 \\
Matched training, s43 & 0.9907 & 0.999 & 0.9917 & 0.0000 & 0.0000 \\
Matched training, s44 & 0.9868 & 0.995 & 0.9888 & 0.0000 & 0.0000 \\
\bottomrule
\end{tabular}}
\end{table}

\section{Physical systems}

Two kinds of world beyond the automaton, each with an exact simulator: billiards, the discrete comparison behind Figure~\ref{fig:billiards} and Section~\ref{sec:stability} (Appendix~\ref{app:billiards}), and three exploratory experiments on continuous dynamics (Appendix~\ref{app:bridge}).

\subsection{Billiards and the Game of Life}
\label{app:billiards}

\begin{figure}[t]
\centering
\includegraphics[width=\linewidth]{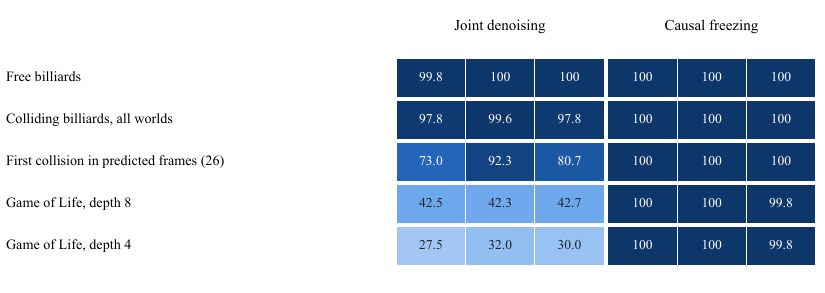}
\caption{\textbf{Same plain denoiser, two samplers, full chains.} SeqAcc (\%), seeds 42/43/44 shown in three separately shaded blocks per cell. Each world has 512 test trajectories, except the first-collision subset of 26 worlds. Joint denoising uses 49 network calls per rollout; causal freezing runs a full 49-level chain for every frame. Depth 8 unless stated. Both samplers read the same weights within a row.}
\label{fig:billiardsfull}
\end{figure}

\paragraph{Setup.}
Billiards live on a $16\times16$ torus: three point balls, each with a constant velocity drawn from the eight king moves. A frame is the occupancy of the balls; a rollout has 8 observed and 8 predicted frames, as in the automaton protocol. In colliding billiards, balls that land on the same cell annihilate. The denoiser is the standard denoiser deepened from four to eight 3D-convolutional layers (196{,}417 parameters), trained with an independent noise level per frame: 200k steps at batch 64 on billiards, and 40k steps at batch 512 on the Game of Life, the training budget of the standard diffusion model; one network per world and seed. Joint denoising shares each of the 49 chain levels across all predicted frames. Causal freezing runs frame $k$'s chain, commits and freezes it, and holds frames not yet started at the top noise level. Both samplers read the same weights in the same evaluation pass; Life worlds come from the $8\times8$ L1 corpus. Figure~\ref{fig:billiards}b scores non-static Life worlds only, whose eight future frames do not all repeat the last observed one. Figure~\ref{fig:billiardsfull} gives per-seed values at the full chain.

\paragraph{Where the joint sampler fails.}
On the Game of Life, the joint frame-perfect rate by predicted frame (full chain, seed mean) is 1.000/\allowbreak 1.000/\allowbreak 0.950/\allowbreak 0.386/\allowbreak 0.335/\allowbreak 0.324/\allowbreak 0.314/\allowbreak 0.314 at depth 4 and 1.000/\allowbreak 1.000/\allowbreak 1.000/\allowbreak 0.987/\allowbreak 0.541/\allowbreak 0.448/\allowbreak 0.441/\allowbreak 0.443 at depth 8; causal freezing stays at or above 0.999 at every frame. Doubling the depth moves the first frame with a rate below 0.90 from frame 4 to frame 5 on every seed, but the sharp drop remains. Trained with one noise level shared by all frames, as the standard diffusion model is, the depth-8 network gives 0.4102/0.3867/0.3828 under joint denoising. On colliding billiards, the joint failures sit in the last one or two predicted frames (frame-perfect 0.999 at frame 7, 0.985 at frame 8), where a single network pass reaches back to only one clean observed frame.

\paragraph{Compute-matched sampling.}
With the same total number of denoiser calls $B\in\{8,16,24,40,49\}$ per rollout for every sampler (causal freezing splits $B$ over its eight chains), freezing stays at 99.8--99.9\% on the Game of Life at every budget (one call per frame suffices) and joint denoising at 42.2--43.2\% (depth 8); joint denoising of free and colliding billiards gives 99.2--99.9\% and 97.2--98.4\%. At $B=8$, the budget of Figure~\ref{fig:billiards}a and Section~\ref{sec:stability}, joint denoising reaches 42.2\% against 99.9\% with freezing on Life, and 99.2\% against 100\% on free billiards.

\paragraph{Unseen rules.}
On unseen rules (L3), the plain denoiser already misses the first predicted frame under joint denoising (frame-perfect 0.17--0.27), and ordering does not make it exact: at depth 4, joint 0.0000 on every seed against 0.1445/0.1519/0.1587 with freezing; at depth 8, 0.1577/0.1655/0.1616 against 0.1411/0.0781/0.1240. Order does not supply the pairing of Section~\ref{sec:temporal}.

\paragraph{Confidence order in plain denoisers.}
A zero-training follow-up lets the depth-8 checkpoints of Figure~\ref{fig:billiardsfull} choose their own commitment order (Table~\ref{tab:plainorder}). It reruns them on CPU in fp32 with sampling seed 0 and carries its own joint and causal-freezing controls, since CPU noise differs from the GPU evaluations of Figure~\ref{fig:billiards}. In each round one network call gives both the committed prediction and the confidence of every pending frame, $c_k=\operatorname{mean}|\tanh(u_k/2)|$ over its cells; pending frames stay at the top noise level. Confidence-first commits the most confident frame (ties to the earliest index), random order a random one, and the threshold sampler every frame with $c_k\ge0.99$, or the most confident one if none qualifies, with the threshold fixed across worlds and seeds.

\begin{table}[h]
\centering
\caption{\textbf{Commitment order in the plain depth-8 denoiser.} SeqAcc (\%) and mean network calls per rollout, all worlds; means over seeds 42/43/44, CPU fp32 reruns of the checkpoints of Figure~\ref{fig:billiardsfull}.}
\label{tab:plainorder}
\small
\adjustbox{max width=\columnwidth}{%
\begin{tabular}{lrrrr}
\toprule
& \multicolumn{2}{c}{Free billiards} & \multicolumn{2}{c}{Game of Life} \\
Sampler & SeqAcc (\%) & Calls & SeqAcc (\%) & Calls \\
\midrule
Joint & 99.4 & 8 & 42.5 & 8 \\
Causal freezing & 100 & 8 & 100 & 8 \\
Confidence-first & 100 & 8 & 100 & 8 \\
Random order & 84.2 & 8 & 66.5 & 8 \\
Threshold parallel & 95.2 & 1.97 & 99.8 & 1.60 \\
\bottomrule
\end{tabular}}
\end{table}

Confidence-first matches causal accuracy on all six checkpoints, within the preregistered two percentage points. This does not isolate a learned ranking: the top confidence is tied in 80.7--87.5\% of Life world-rounds and 87.5\% of billiard world-rounds, often because the fp32 confidence saturates, and ties go to the earliest frame. The preregistered parallelism prediction is not met on accuracy or on rounds: billiard accuracy falls by more than two points on every seed, and Life takes fewer rounds than billiards (1.60 against 1.97 calls), not the two additional rounds predicted. On non-static Life worlds alone, parallel commitment still reaches 99.8\% in 1.91 calls. These weights can thus commit several Life frames per call reliably under this schedule; the results support neither a universal one-frame-at-a-time requirement nor an automatic choice of parallelism matched to the dynamics.

\subsection{Exploratory experiments on continuous dynamics}
\label{app:bridge}
\providecommand{\capitem}[1]{\par\hangindent=0.9em\hangafter=1\noindent\makebox[0.9em][l]{\textbullet}\textbf{#1}~}

\begin{figure}[t]
\centering
\includegraphics[width=\linewidth]{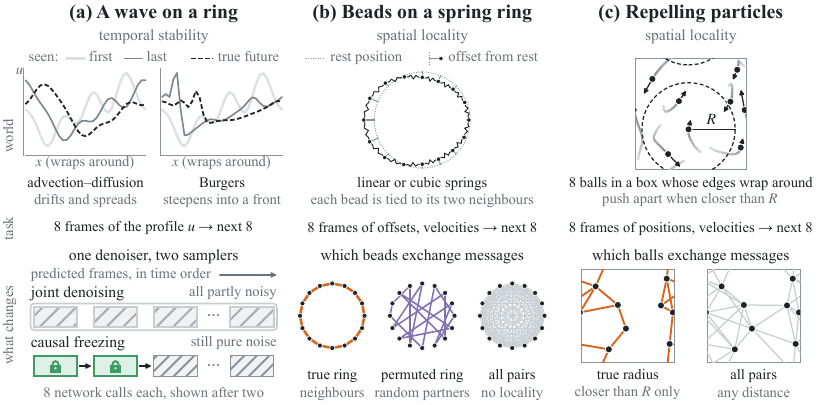}
\caption[The three continuous worlds]{\textbf{The three continuous worlds, what the network sees, and what each comparison changes.} Pictures use the first test episode, not selected by outcome; colours match Figure~\ref{fig:bridgebars}.
\capitem{(a) Wave on a ring.} The profile $u$ at the first and last seen frames and the true future at the last predicted frame, for advection--diffusion and Burgers ($x$ wraps around). Below, one trained denoiser with two samplers, 8 network calls each, sketched after two calls: joint denoising refines all predicted frames together, so all are partly noisy; causal freezing spends one call per frame and freezes it (green) before starting the next, which is still pure noise.
\capitem{(b) Beads on a spring ring.} Beads at the last seen frame, moved off the rest circle in proportion to their offset, each tied to its two neighbours. Below, the pairs that may exchange messages: the true ring (orange), a fixed permuted ring with the same number of partners per bead (purple; it keeps one true link), or all pairs (grey).
\capitem{(c) Repelling particles.} Paths over the 8 seen frames and velocities at the last; the dashed circle has the interaction radius $R$ and, like the paths, wraps around the box edges. Below, messages only between balls closer than $R$ (orange; distances measured across the wrapped edges) or between all pairs (grey).}
\label{fig:bridgetasks}
\end{figure}

Physical systems keep what makes the automaton useful---a simulator computes the correct future---but replace binary cells with continuous states. Three small exploratory experiments ask whether two of the three properties carry over (Figure~\ref{fig:bridgetasks}). As in the main text, each comparison changes one thing, holds the rest of the model fixed, and reports every planned arm and seed (42/43/44).
\begin{itemize}
\item \textbf{A wave on a ring} (temporal stability). A profile $u$ on 32 periodic points drifts and spreads (advection--diffusion, $u_t=-a\,u_x+\nu\,u_{xx}$) or also steepens into a front (viscous Burgers, adding $-u\,u_x$). The network sees 8 frames of the whole profile and predicts the next 8. \emph{What changes:} only how one trained denoiser is sampled---all 8 frames refined together (joint denoising) or one frame per call, each settled before the next (causal freezing)---with the same weights, the same initial noise, and 8 calls.
\item \textbf{Beads on a spring ring} (spatial locality). Sixteen beads sit on a ring, each tied to its two neighbours by a damped spring that is linear or cubic. The network sees every bead's displacement and velocity for 8 frames and predicts the next 8. \emph{What changes:} which beads may exchange messages in one message-passing network---the true neighbours, a permuted ring with as many links (it keeps 1 of the 16 true links), or all pairs.
\item \textbf{Repelling particles} (spatial locality). Eight damped balls in a periodic square push each other apart when closer than a radius $R$. The network sees their positions and velocities for 8 frames and predicts the next 8. \emph{What changes:} whether messages flow only between balls closer than $R$, or between all pairs.
\end{itemize}
In every world the coefficients (flow speed, viscosity, stiffness, force strength, damping) vary between episodes and are never shown to the network. Test coefficients lie inside the training range (two spring damping values lie just above it), so these experiments test interpolation within a known equation family, not a new law. Each experiment has its own normalized error (Table~\ref{tab:bridge}); errors compare arms within an experiment, not across experiments or with SeqAcc. A fourth experiment, on temporal locality, gave the wave predictor adjacent-frame differences instead of repeated frames: a small mean gain on advection--diffusion (two of three seeds) and none on Burgers (one of three); it is not detailed here. All runs and audits are registered with the other results.

\begin{figure}[t]
\centering
\includegraphics[width=\linewidth]{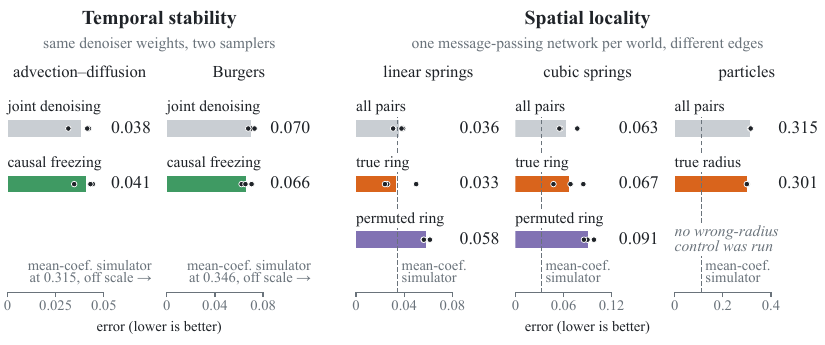}
\caption[Continuous dynamics: test error of each arm]{\textbf{Test error of every arm; lower is better.}
\capitem{Bars and numbers:} means over seeds 42/43/44; dots are single seeds.
\capitem{Colours:} grey, the standard arm; green, causal freezing; orange, the true interaction structure; purple, the permuted ring. The coloured arm is not always the better one.
\capitem{Reference:} the dashed line (or off-scale note) is a simulator run with the mean training coefficients; it needs no learning but knows the equation.
\capitem{Scales:} each panel has its own axis and error measure (Table~\ref{tab:bridge}); compare bars within a panel only.
\capitem{Particles:} two arms only; no wrong-radius control was run.}
\label{fig:bridgebars}
\end{figure}

\begin{table}[h]
\centering
\small
\caption[Continuous dynamics: all arms]{\textbf{Continuous dynamics: all arms} (test split; mean $\pm$ sample SD over seeds 42/43/44; lower is better).
\capitem{Seeds won:} seeds on which the arm listed second beats the arm listed first.
\capitem{Wave:} global normalized RMSE over the 8 predicted frames; final weights.
\capitem{Springs:} RMSE at the 8th predicted frame, each channel scaled by its training SD; final weights.
\capitem{Particles:} RMSE with fixed physical scales, averaged over the 8 predicted frames; checkpoint chosen on validation (at 500 or 1{,}000 of 2{,}000 updates).}
\label{tab:bridge}
\adjustbox{max width=\columnwidth}{%
\begin{tabular}{lllc}
\toprule
World & Arms & Error & Seeds won \\
\midrule
\multicolumn{4}{l}{\emph{Temporal stability: the same denoiser weights under two samplers, eight calls each}} \\
Advection--diffusion & joint / causal freezing & $0.0384\pm0.0059$ / $0.0407\pm0.0053$ & 0/3 \\
Burgers & joint / causal freezing & $0.0702\pm0.0026$ / $0.0660\pm0.0042$ & 3/3 \\
\midrule
\multicolumn{4}{l}{\emph{Spatial locality: supplied interaction structure}} \\
Linear spring ring & permuted ring / true ring & $0.0579\pm0.0030$ / $0.0333\pm0.0145$ & 3/3 \\
 & \quad all pairs & $0.0359\pm0.0045$ & \\
Cubic spring ring & permuted ring / true ring & $0.0910\pm0.0064$ / $0.0669\pm0.0187$ & 3/3 \\
 & \quad all pairs & $0.0627\pm0.0125$ & \\
Repelling particles & all pairs / true radius & $0.3145\pm0.0015$ / $0.3006\pm0.0013$ & 3/3 \\
\bottomrule
\end{tabular}}
\end{table}

\paragraph{Temporal stability: the better sampler depends on the dynamics.}
Joint denoising is better on advection--diffusion and causal freezing on Burgers, on all three seeds each (Figure~\ref{fig:bridgebars}). This loosely echoes billiards against the Game of Life (Section~\ref{sec:stability}), where freezing left billiards essentially unchanged and rescued Life, at a far smaller scale: about 6\% of the error in either direction. On Burgers, joint denoising is better at the first predicted frame on every seed, and the advantage of freezing is concentrated in the last three frames, where it holds on every seed (seed-mean error at the eighth frame 0.135 joint against 0.125 frozen). Why each family prefers its sampler was not tested, and the samplers also differ in the noise each frame sees, so the gain is not attributed to commitment alone. The denoiser (one per family and seed) is a 2-D convolution over time and space with 37k parameters, trained for 2{,}000 updates; it reads the eight observed frames at every predicted position.

\paragraph{Spatial locality: the interaction structure matters.}
Hiding the true neighbours hurts: the permuted ring is worse than the true ring on every seed of both spring families. Restricting messages to exactly the true neighbours matters less. All pairs, which include the true neighbours, are comparable to the true ring (the true ring wins two of three seeds in each family; all pairs has the lower cubic mean), and on the particles true-radius messages beat all pairs on every seed, by about 4\%. The spring network has 6.9k parameters and 2{,}500 updates, the particle network 27k parameters and 2{,}000 updates. Both experiments supply the graph or radius and the integration step, and a simulator run with the mean training coefficients is comparable on the linear springs and more accurate elsewhere (dashed lines in Figure~\ref{fig:bridgebars}), so neither experiment shows that the network has learned the law. As with pixel accuracy on the automaton, lower mean error is not rollout reliability: 84\%, 87\%, and 41\% of linear-spring rollouts (true ring, all pairs, permuted ring) stay within 5\% at every predicted frame, and no particle rollout does.

\section{Extended outlook}
\label{app:outlook}

These directions extend Section~\ref{sec:outlook}. They are hypotheses from the automaton experiments, not results.

\paragraph{Physical video generation.}
A temporal window should span one frame more than the order of the dynamics: two frames pair a situation with its outcome in a first-order automaton, while Newtonian motion needs three, since a single frame carries no velocity. An index shift transfers only with motion-compensated alignment, because the entity a key must meet no longer sits at the same pixel one frame later. For the latent-space predictor of Section~\ref{sec:outlook}, these conditions apply to encoded states; a first check is whether the encoder keeps each cell at the same latent position across frames. Generation schedules should be tested against the dependencies a model has learned: freezing may help where it supplies reliable intermediate states, while some futures can be generated jointly. In the structured denoiser, confidence-first commitment recovered frame order, but under the directional priors its architecture already supplies (Appendix~\ref{app:e12}); in the plain denoiser, committing several frames per call kept Life accurate but cost accuracy on free billiards, contrary to the preregistered prediction (Appendix~\ref{app:billiards}), so how far commitment can be parallelized remains open.

\paragraph{Dependencies in diffusion language models.}
Reliable intermediate states may also matter when generated tokens carry an execution, as in program traces and mathematical derivations, where later predictions depend on earlier computed values and jointly revising them may waste calls or destabilize continuation. A direct test holds the weights fixed, compares joint refinement, sequential commitment, and partial commitment at matched call budgets, and measures exact execution. The question concerns computational dependencies, not a blanket division between writing and reasoning.

\paragraph{Running the audit on real video models.}
Worlds with a written-down rule (automata, constant-velocity and colliding particles) can be rendered as video to probe pretrained models with weights fixed: sampler swaps, pairing at the tokenizer level, and strict exact-execution metrics in place of quality scores. Two criteria carry over. Identifiability from the window, the analogue of our self-consistent test corpora, asks whether a model's context window holds enough transitions to pin down the governing law \citep{ha2018world, openai2024sora, bruce2024genie, valevski2024diffusion}. The tolerance--horizon criterion, the analogue of strict SeqAcc, asks over what horizon and at what per-step tolerance a model must stay exact to count as executing the law rather than approximating it; benchmarks already report visual realism outrunning physical understanding \citep{motamed2025physicsiq, bansal2024videophy, bear2021physion}. Physics-structured learners show what building in the right structure buys \citep{sanchezgonzalez2020learning, greydanus2019hamiltonian, cranmer2020lagrangian, chen2018neural, gillman2025force}, and in transformers trained on orbital data the context length steers which world model forms \citep{liu2026newton}.

\paragraph{A testbed for interpretability.}
Every intermediate feature here has a known ground truth (neighbour counts, the 18 situations, the rule bits), so probing and dictionary-learning methods can be scored against the features a model must compute rather than against features found post hoc \citep{elhage2022toy, bricken2023towards}.

\end{document}